\documentclass[lettersize,journal]{IEEEtran}
\usepackage{amsmath,amsfonts}
\usepackage{algorithm}
\usepackage[caption=false,font=normalsize,labelfont=sf,textfont=sf]{subfig}
\usepackage{url}
\usepackage{graphicx}
\usepackage{cite}
\usepackage{booktabs}
\usepackage[bookmarks=true]{hyperref}
\usepackage{mathtools}
\usepackage{afterpage}
\usepackage{bm}
\usepackage{flushend}

\newcommand{\xv}{\vec{x}} %
\newcommand{\Xv}{\vec{X}} %
\newcommand{\gv}{\vec{g}} %
\newcommand{\rv}{\vec{r}} %
\newcommand{\yv}{\vec{y}} %
\newcommand{\vv}{\vec{v}} %
\newcommand{\ev}{\vec{e}} %
\newcommand{\mv}{\vec{m}} %

\renewcommand{\ij}{\overrightarrow{ij}}

\newcommand{\rdgnn}{RD-GNN}
\newcommand{\glatent}{\Xv}
\newcommand{\glatentseg}{\xv}
\newcommand{\glatentprime}{\Xv^{^\prime}}

\renewcommand{\vec}{\mathbf}

\newcommand{\tpar}[1]{\vspace{2pt}\noindent\textbf{#1}}

\newcommand{\actionenc}{\phi_{A}}

\newcommand{\ginput}{\xv^{I}}

\newcommand{\etedyn}{eRDTransformer}
\newcommand{\erdt}{eRDT-trans-dec}
\newcommand{\etemdy}{eRDT-mlp-dyn}

\newcommand{\egnn}{eRD-GNN}

\usepackage[skip=0pt,font=small,labelfont=bf]{caption}
\usepackage{amsmath}
\DeclareMathOperator*{\argmax}{arg\,max}

\newcommand{\shrinka}{\def\baselinestretch{0.993}\large\normalsize}
\usepackage{color}

\begin{document}

\title{Latent Space Planning for Multi-Object Manipulation with Environment-Aware \\ Relational Classifiers}

\author{Yixuan Huang$^{1}$, Nichols Crawford Taylor$^{1}$, Adam Conkey$^{2}$, Weiyu Liu$^{3}$, Tucker Hermans$^{1, 4}$%
  \thanks{$^1$~University of Utah Robotics Center and Kahlert School of Computing,
  University of Utah, Salt Lake City, UT, USA. $^2$ HRL Laboratories, Malibu, CA. $^3$ Georgia Tech, GA, USA. 
  $^4$ NVIDIA Corporation, USA.  yixuan.huang@utah.edu, 
  nichols.crawfordtaylor@utah.edu,
  ajconkey@hrl.com, 
  wliu88@gatech.edu, 
  thermans@cs.utah.edu}}

\input{overview_fig}
\maketitle

\pagestyle{headings}
\makeatother

\begin{abstract}
Objects rarely sit in isolation in everyday human environments.
If we want robots to operate and perform tasks in our human environments, they must understand how the objects they manipulate will interact with structural elements of the environment for all but the simplest of tasks.
As such, we'd like our robots to reason about how multiple objects and environmental elements relate to one another and how those relations may change as the robot interacts with the world.
We examine the problem of predicting inter-object and object-environment relations between previously unseen objects and novel environments purely from partial-view point clouds. 
Our approach enables robots to plan and execute sequences to complete multi-object manipulation tasks defined from logical relations. 
This removes the burden of providing explicit, continuous object states as goals to the robot.
We explore several different neural network architectures for this task.
We find the best performing model to be a novel transformer-based neural network that both predicts object-environment relations and learns a latent-space dynamics function. We achieve reliable sim-to-real transfer without any fine-tuning. Our experiments show that our model understands how changes in observed environmental geometry relate to semantic relations between objects.  We show more videos on our  
website: \url{https://sites.google.com/view/erelationaldynamics}.

\end{abstract}

\begin{IEEEkeywords}
Multi-object manipulation, Learning for motion planning,  Semantic manipulation.
\end{IEEEkeywords}

\section{Introduction}\label{sec:intro}
\IEEEPARstart{F}{or} robots to act as our assistants, caregivers, and coworkers at home, at work, and in the wild they will need to contend with varying environments.
Environmental structure has profound impacts on manipulation. 
Placing a large tray on a small table is much different than placing a cup onto a high shelf.
The task and motion planning~(TAMP) formulation shows promise in addressing tasks of this nature~\cite{kim2019learning, garrett2017sample, kim2020learning, driess2020deep, garrett2020pddlstream, garrett2020integrated, liang-icra2022, curtis2022long}. 
However, TAMP solvers typically assume known relational dynamics (e.g. symbolic/logical effects of actions)~\cite{curtis2022long, garrett2020pddlstream, garrett2017sample, garrett2020integrated, kim2019learning, driess2020deep} or they assume known object pose and access to explicit 3D object models~\cite{liang-icra2022, garrett2020pddlstream, garrett2017sample, garrett2020integrated, kim2019learning}. 
No work to date explicitly reasons about object-environment relational dynamics with unknown objects observed from partial-view data.

Robots operating in varying human environments must contend with many objects at a time. 
This could include future home assistants putting away groceries, warehouse robots packaging items for shipment, or grocery store robots stocking shelves.
As such, robotic multi-object manipulation and rearrangement has received much attention in the literature~\cite{cosgun-iros2011,chang2012interactive,gupta2012using, panda2013learning,dogar2014object,murali20206,paxton-corl2021-semantic-placement,liu-icra2022-structformer,zhu2020hierarchical, pan2022algorithms}. The most recent of these works show excellent results in reasoning about novel objects from partial view sensory information~\cite{murali20206,paxton-corl2021-semantic-placement,liu-icra2022-structformer,zhu2020hierarchical,li2020towards, lin2022efficient, sharma2020relational}.
However, robots using these approaches operate in a limited capacity manipulating individual objects one at a time.
In contrast, some works have shown the ability for robots to reason about and manipulate multiple objects at once~\cite{wilson2020learning,suh2020surprising}. These works leverage image-based feedback controllers, that lack the level of semantic reasoning and explicit object grounding we desire for multi-step planning to logical goals.

Research over the past few years has shown increasing ability for robots to manipulate and rearrange novel objects in cluttered environments by learning to predict the effects of the robot's actions~\cite{murali20206,paxton-corl2021-semantic-placement,liu-icra2022-structformer,zhu2020hierarchical, lin2022efficient, sharma2020relational}.
Significantly less attention has been given to reasoning about novel environments and how environmental structures relate and interact with the objects being manipulated.
For example, a robot should be able to reason that an object on a high shelf is above an object in a drawer lower in the same cabinet. While traditional planning-based manipulation approaches reason about the environment, these are typically monolithic geometric representations focusing on collision avoidance for planning and navigation~\cite{hornung13auro,kuntz-isrr2017-anytime-planning,Place-SaxenaPlaceLearn2012,paxton-corl2021-semantic-placement,cosgun-iros2011}.
However, these models of immobile environmental components have not been fully integrated into neural representations for multi-object manipulation. In particular we desire that the robot can reason about logic-based object-environment goals to enable efficient task planning~\cite{garrett2020pddlstream}. 

We further advocate for the use of logical relations for specifying goals as in~\cite{paxton-corl2021-semantic-placement,zhu2020hierarchical} as they provide a useful language for communication between robot and human. 
A human can easily construct a goal for tasking a robot by providing a conjunction of desired logical relations between objects in the scene. On the flip side, the robot can use its predictions of logical relations to communicate its belief over the current scene or future states it intends to achieve through manipulation. 
This contrasts with several recent approaches to rearrangement which provide images as goals to the robot~\cite{Qureshi-RSS-21}. 
Generating images for all desired goals requires a much higher burden on the user and in many cases would require the user to actively rearrange the scene, obviating the need for the robot.

To enable reasoning about manipulation effects simultaneously on multiple objects and environments, we propose a learned relational dynamics model with a structured, object-aware latent state space for use in TAMP. 
Motivated by recent success in robotics and vision problems we choose to examine both transformers-based neural network architectures~\cite{vaswani2017attention,brohan2022rt, wang2022generalizable} and graph neural networks~\cite{graph_nets,sharma2020relational, simeonov2020long}.
A core technical question we address is how to integrate these multiple distinct environmental components and objects into a single learnable model.

We contribute two novel models to TAMP, \emph{environment-aware} relational dynamics transformer (\etedyn{}) and relational dynamics graph neural network (\rdgnn{}). Compared to traditional TAMP approaches, the main novelties of our approaches are (1): our proposed approaches take as input a segmented, partial view point cloud of the objects and environments in the scene. 
(2): Both models encode this observation into the latent space, from which the network can predict inter-object and object-environment relations for both the current scene and future states given a sequence of actions. To enable these future predictions, we learn a dynamics function (i.e. state transition function) in the latent space. 
We then use the learned network to perform planning to achieve a desired relational goal.

Fig.~1 illustrates how our learned model can be used for multi-step planning. 
Importantly, our planning framework can incorporate multiple distinct robot skills and produce multi-step task plans to achieve the specified goals.
Our planner performs diverse multi-object rearrangements including picking and placing multiple objects to different layers of a bookshelf, pushing objects to the boundary or off a table, pushing objects to be in contact on a bookshelf, and lifting and placing multiple objects at once. 

We show that \etedyn{} and \rdgnn{} outperform similarly structured multi-layer perceptrons (MLPs) operating directly on object pairs both in terms of planning success rate and predicting post-manipulation relations to enable successful planning. 
Crucially, using a transformer or graph neural network allows the robot to use the same model to reason about a variable number of objects. Further, by directly using partial view point cloud information as input, the robot can reason about objects of novel shape and size without access to explicit object models.

We perform extensive simulated and real world experiments to further test the hypothesis that we can train \etedyn{} and \rdgnn{} using only the pre- and post-manipulation relational labels for supervision, in addition to the input point cloud and actions. 
Our experiments show that using this relational supervision outperforms training to predict changes in object pose, coupled with an analytic approach to predicting relations from the object bounding boxes. Further, we show that training with both relational and pose estimation losses provides no real benefit over training with relations alone.

This manuscript significantly extends previous work using \rdgnn{} presented in~\cite{Huang-icra2023-graph-relations}.
The specific extensions include: (i) We demonstrate that planning using the learned network enables a robot to achieve many multi-object rearrangement tasks in different environments.
We evaluate these tasks on a variety of different furniture shapes and styles showing the ability of our network to understand how relations change as a function of both object and environmental geometry. We additionally show that our model trained purely in simulation transfers reliably to the real world with no fine-tuning; and (ii) we compare \etedyn{} to both \rdgnn{} and an extension of the graph-based neural networks (GNN) model \egnn{}. We show \etedyn{} performs much better compared to \rdgnn{} and \egnn{} in both simulation and real world evaluation.
Through ablation we find that using a transformer to encode the latent-space dynamics offers the biggest benefit over alternative models. Finally, (iii) we propose learning a classifier to predict whether a specific segment in the point clouds is a movable object or fixed part of the environment (e.g. a shelf). We use this classifier as a skill precondition and verify that its use improves the effectiveness of task planning.

\section{Related Work}\label{sec:related-work}
Neural networks, including graph neural networks, have been applied to reason about spatial relationships and perform planning based on said reasoning~\cite{simeonov2020long,zhu2020hierarchical, paxton-corl2021-semantic-placement, liu-icra2022-structformer,yuan2022sornet}.
Paxton et al.~\cite{paxton-corl2021-semantic-placement} present a framework to reason about pairwise relations and plan to find an object placement that is physically feasible and satisfies the goal relations. 
The pairwise, MLP-based model allows only one object to be manipulated at a time by considering rigid object transformation via pick-and-place. In contrast, we examine GNNs and transformers to handle multi-object interactions where more than one object may move during a single action.
Zhu et al.~\cite{zhu2020hierarchical} presents a grounded hierarchical planning framework for long-horizon planning manipulation tasks that leverages a symbolic scene graph to predict high-level plan actions and a geometric scene graph to predict low-level motions. Unlike our work, Zhu et al.~\cite{zhu2020hierarchical} do not examine multi-object dynamic interactions.
Lou et al.~\cite{lou2022learning} predict spatial relations between objects in clutter using GNNs to aid in finding better grasps, but do not model how relations will change post grasp. Furthermore, Driess et al.~\cite{driess2022learning} learn to predict multi-object interactions using graph nets, with supervised reconstruction for NERF-like embeddings. Unlike our proposed approach, they do not predict object relations and learn and plan at a much finer time scale which makes their simulation-only experiments unlikely to transfer well to the real world. Biza et al.~\cite{biza2022factored} similarly examine learning object-oriented models of the world with pose estimation supervision. They show the ability to embed pose-based goals into a latent space, but do not explicitly reason about relations or manipulating multiple objects at once.
In~\cite{lin2022efficient} a GNN-based policy learns to perform multi-object rearrangement tasks including stacking and unstacking. However, the policy requires full object pose information and manipulates one object at a time.

\begin{figure*}[th]
    \centering
    \includegraphics[width=2\columnwidth]{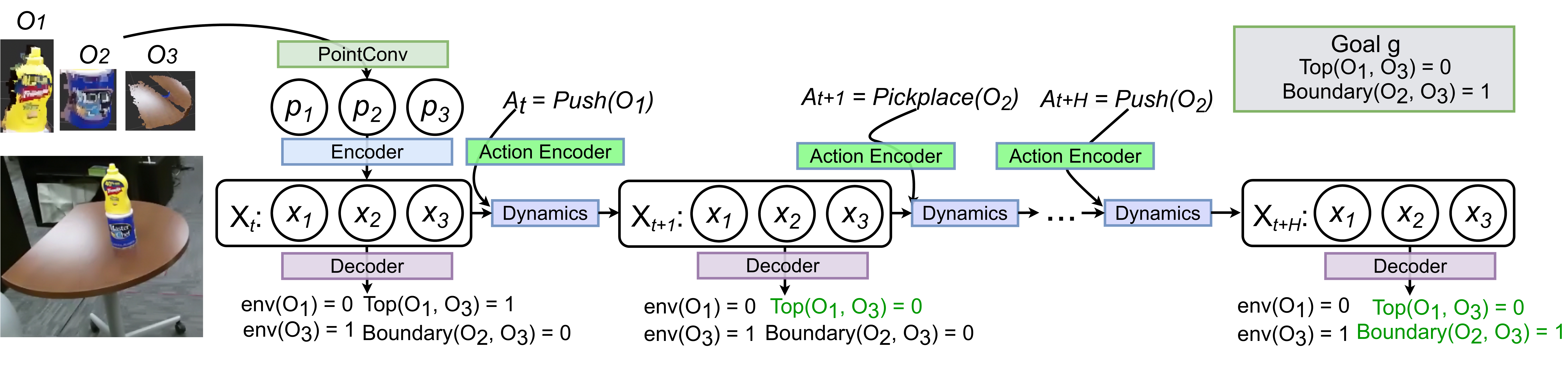}
    \caption{Taking a segmented, partial-view point cloud as input, we first process it using PointConv to generate segment-specific features. We then pass these features into an encoder to predict a latent state $\glatent$. We can decode $\glatent$ to predict both if the segment is a movable object and relations between objects and environment segments. By learning an action-conditioned latent-space dynamic model, our approach can be used to solve multi-step planning problems. In green we highlight relations that satisfy relations in the logical goal \(\gv\).}
    \label{fig:approach}
\end{figure*}

Transformer-based approaches~\cite{vaswani2017attention, brohan2022rt} have shown impressive results in robot manipulation research.
Brohan et al.~\cite{brohan2022rt} propose a ``Robotics Transformer'' system to perform many real world manipulation tasks but it requires a huge number of demonstrations.
Liu et al.~\cite{liu-icra2022-structformer, liu2022structdiffusion} leverage transformer and diffusion models to rearrange objects based on language goals but they can only move one object each time.
Wang et al.~\cite{wang2022generalizable} leverage transformer model and pretrained scene representation to enable multi-step manipulations in household environments. However, they are limited in 2D space since they use pixel location to do the planning which might limit their model's capability in real world applications. Our work focuses on multi-step manipulation in 3D space with strong results in the real world.
Zhu et al.~\cite{zhu2022viola} leverage a transformer to improve the performance of imitation learning approach for robot manipulation. However, they do not reason about the relations between objects and environments and are limited in 2D space. 
Liu et al.~\cite{liu2022instruction} pretrains a multimodal transformer to encode vision observations and language instructions. They train a transformer policy to perform different manipulation tasks.
Furthermore, each of \cite{liu2022instruction, liu-icra2022-structformer, liu2022structdiffusion, brohan2022rt} consider natural language. Our proposed framework is a variant of TAMP leveraging logical relations to enable efficient task planning. We view this as a benefit compared to natural language in terms of structuring the search and offering easier human inspection of the model and plans.

Reasoning about the environment during robot manipulation is important since objects always interact with varied environments.
Garrett et al.~\cite{garrett2020online} propose an online learning approach to enable multi-step manipulation in the kitchen environment with only partial observable information.
Kase et al.~\cite{kase2020transferable} proposes a hybrid approach to learn both high-level symbolic plans and low-level motion planner in a photorealistic kitchen scenario.
Radosavovic et al.~\cite{radosavovic2022real} leverages the visual pretraining to build useful embeddings for real world robotics task with different environments.
Silver et al.~\cite{silver2022learning} learn neuro-symbolic skills for bilevel planning and show several experiments with different environments.
Wada et al.~\cite{Wada_ICRA2022_reorient} achieved nice results about reorienting objects into different layers of bookshelf but they require object models and can only move one object each time.
Kim et al.~\cite{kim2022graphdistnet} present a graph-based approach to predict collision between a robot and the surrounding environment.
Agia et al~\cite{agia2022taps} propose a framework to coordinate different action primitives that can achieve long-horizon planning with different environments and skills. However, the approach only operates on low-dimensional state features like object pose.

Traditionally 3D environment models built from sensor data provide a monolithic geometric representation focusing on collision avoidance for planning and navigation~\cite{hornung13auro,kuntz-isrr2017-anytime-planning} or object placement~\cite{Place-SaxenaPlaceLearn2012,paxton-corl2021-semantic-placement,cosgun-iros2011}.
Other works examine building semantic segmentation maps of environments often using multiple sensor readings over time~\cite{RUSU2008927, khandelwal2021segmentation} in order to locate and possibly manipulate objects~\cite{Wada:etal:CVPR2020,Wada:etal:ICRA2022a,Sucar:etal:3DV2020} or fixtures in the environment~\cite{garrett2020online}. These methods often rely on accurate models of the objects being manipulated~\cite{garrett2020online, Wada_ICRA2022_reorient,Wada:etal:CVPR2020} or don't explicitly reason about the object-environment relations or how those could change through robot intervention~\cite{liu2022instruction, shridhar2022peract, zhu2022viola}.

Long-horizon planning has become an important problem for robot manipulation.
Task and motion planning (TAMP)~\cite{kim2019learning, garrett2017sample, kim2020learning, driess2020deep, garrett2020pddlstream, garrett2020integrated, liang-icra2022} defines a promising method to solve long horizon problems. TAMP approaches typically assume models of how objects and potentially their relations change. While learning has been used for various aspects of TAMP, no work has shown how to plan with multi-object dynamic interactions from point cloud data.
Simeonov et al.~\cite{simeonov2020long} propose an approach to object manipulations from point cloud data. They leverage a plan skeleton to solve long horizon planning problems. 
However, they do not reason about object relations and only manipulate one object with each action. 
Furthermore, they assume multiple cameras to get effectively full point clouds of each object, while we use partial-view point clouds obtained from a single camera.
Liang et al.~\cite{liang-icra2022} learn to plan with different skill primitives which sometimes include multi-object dynamic interactions. They perform multi-step skill planning using a heuristic graph search. However, they assume knowledge of object state and do not explicitly reason about object relations for learning.
Curtis et al.~\cite{curtis2022long} learn perception modules to rearrange unknown objects into a goal region. However, they assume the effects of actions are known and do not reason about object-environment relations. Furthermore, they rely on shape completion to reconstruct object meshes from point clouds while we directly plan with partial-view point clouds.

\section{Preliminary Knowledge}
\subsection{Graph Neural Networks}
We define a directed graph $G = \{V, E\}$ with nodes $V = \{\vv_i\}$ and edges as \(E = \{\ev_{\ij}\}\) where each \(\vv_i\) and \(\ev_{\ij}\) is a feature vector for node \(i\) or the edge from \(i\) to \(j\) respectively. We seek to encode information associated with this graph into a neural network; following~\cite{graph_nets} we can reason about our graph network operations in terms  of message passing in the graph, where a single graph net layer of \emph{update} and \emph{aggregation} functions performs one round of message passing between neighbors in the graph. By constructing multiple graph layers, information from nodes across the graph can propagate in the form of deeper and deeper features. %

Update functions transform individual node or edge features. We use feed-forward multi-layer perceptrons as update functions in this paper. We denote node updates as $\vv_{i}^{\prime} = f_{n}(\vv_{i})$ and edge updates as $\ev_{\ij}^{\prime} = f_{e}(\ev_{\ij})$. Aggregations take inputs from multiple parts of the graph and reduce them to a fixed feature length, thus enabling consistent output feature dimensions from a variable input size. We denote a message from node \(i\) to node \(j\) as \(\mv_{\ij} = (\vv_i \oplus \vv_j \oplus \ev_{\ij})\) and define our message update functions as \(\mv_{\ij}^\prime = f_{m}(\mv_{\ij})\). Here \(\oplus\) denotes vector concatenation. To define our aggregation functions we introduce an intermediate variable \(\yv_i = \frac{1}{|\mathcal{N}(i)|}\sum_{j \in \mathcal{N}(i)}(\mv_{ji}^{\prime})\) which takes the average of all messages incoming to node \(\vv_i\) denoted as those coming from nodes in node \(i\)'s neighborhood \(\mathcal{N}(i)\).
Using this we can define our node aggregation function as $\vv_{i}^{\prime} = g_v(\vv_{i} \oplus \yv_i)$ and the edge  aggregations as \(\ev_{\ij}^{\prime} = g_{e}(\vv_i \oplus \vv_j \oplus \yv_{i} \oplus \yv_{j} \oplus \ev_{\ij})\)
where \(g_v(\cdot)\) and \(g_{e}(\cdot)\) define MLPs.
This edge aggregation thus concatenates and then transforms the features associated with the two neighboring nodes and the messages passing between them.
For more details on graph nets including alternative aggregation functions see~\cite{graph_nets}.

\subsection{Transformer Neural Networks}
Transformers are an attention-based approach for sequential data.
The input for a transformer contains a sequence of length \(n\), which we denote $N = {n_1, ..., n_{n}}$ and outputs node features as $N^{\prime} = n_1^{\prime}, ..., n_{n}^{\prime}$.
The core part of the transformer is to sequentially get the attention between different nodes, where we view this interaction as a special case of message passing.
Specifically, each input feature $n_i$ will be linearly projected to a query $q_i$, key $k_i$, and value $v_i$. Then each output $n_{i}^{\prime}$ is a weighted sum of values based on attention.
The function to compute attention is $\texttt{Attention}(Q,K,V) = \texttt{softmax}(\frac{QK^{T}}{\sqrt{d_{k}}})V$, where $d_k$ is the dimension of keys $K$.
We can pass the input $N$ through $k$ layers of transformer to get useful output sequences of varying lengths.
For more details on transformers, we refer you to the original paper~\cite{vaswani2017attention}.

\section{Problem Definition}
We assume the robot perceives the world as a point cloud $Z$ with \(N\) associated object and environment segments $O_{i} \subset Z, i = 1,2,...,N$.
The robot receives a goal defined as a logical conjunction of \(M\) desired object and environment relations,  $\gv = r_{1} \land r_{2} \land ... \land r_{M},  r_j \in \mathcal{R}$, where
$\gv$ represents the goal conjunction, $r_j$ denotes each goal relation, and $\mathcal{R}$ represents the set of all possible relations.
Example relations in $\mathcal{R}$ include planar spatial relations such as ``entity \textit{i} is in front of entity \textit{j}'' or 3D relations such as ``entity \textit{i} is on top of entity \textit{j}'' and ``entity \textit{i} is in contact with entity \textit{j}.''
Interestingly, since we have the environment as part of the relations, we can define the goal relations like \(\texttt{Above}(O_i, O_k)\) where \(O_k\) defines an environment segment such as a table or shelf.

We provide our robot with a set of \(L\) parametric action primitives $\mathcal{A} = \{A_1, \ldots, A_L\}$ where \(A_l\) defines the discrete skill, which has associated skill parameters $\theta_{l}$. For example, a push skill, \(A_l\), with parameter, $\theta_{l}$, encoding the end effector pose and push length or a pick-and-place skill defined by the grasp and placement poses.

We define the robot's planning task as finding a sequence of skills and skill parameters $\tau = (A_{0}, \ldots, A_{H-1})$ that when sequentially executed transform the objects such that they satisfy all relations defined in the logical goal $\gv$.

We can now formally define our planning objective as maximizing the probability of achieving the goal relations with the following constrained optimization problem:
\begin{align}
  \argmax_{\tau=(A_{0}, \ldots, A_{H-1})} &\prod_{k=1}^{H} P(\rv_{k} = \gv_{k} | \xv_{k}) = D_r(\xv_{k}) \label{eq:planning-obj} \\
  \texttt{subject to} \;\; &\xv_{k+1} = \delta(\xv_k, A_k) \; \forall k = 0, \ldots, H-1 \label{eq:dynamics-constraint}\\
               &\xv_{0} = E(E_p(Z_0))  \label{eq:latent-grounding}\\
               &A_k \in \mathcal{A}  \;\;\;\;\;\;\;\;\;\;\;\;\;\;\;\,\forall k = 0, \ldots, H-1 \label{eq:skill-domain} \\ %
               &\theta_{min} \preceq \theta_k \preceq \theta_{max} \;\forall k = 0, \ldots, H-1 \label{eq:skill-constraints}
\end{align}
The constraints in this optimization problem encode the latent space dynamics (Eq.~(\ref{eq:dynamics-constraint})), grounding of the initial latent state from the observed point cloud Eq.~(\ref{eq:latent-grounding}), and constraints on the action parameters Eqs.~(\ref{eq:skill-domain}--\ref{eq:skill-constraints}).
We thus chain together predicted action effects decoding each state to predict the inter-object relations. Fig.~1 visualizes planning with this model.

After solving this optimization problem the robot can execute the planned actions in the physical world. Our proposed network enables the robot to validate if it achieved its goal by computing \(\rv_k = D_r(E(E_p(Z_k)))\). Where \(Z_k\) denotes the current point cloud observation.

\section{Learning Multi-Object and \\
Object-Environment Relational Dynamics}\label{sec:approach}

\label{high_level_approach}
We propose using a latent-space approach to planning.
We visualize our approach and the components of the neural network model in Fig~\ref{fig:approach}. The learned model contains three main components: an encoder, decoder, and latent dynamics model. We now provide an overview of the input-output structure for each component and a brief overview of how we can use it for planning. We give details on the specific instances we implement of each component, as well as how we train the model in subsequent sections.

\tpar{Encoder:} The model takes in a segmented point cloud of the current observation $Z_t$. We pass the point cloud segments into our point cloud encoder~\cite{wu2019pointconv} to get a feature vector for each segment, $P_{i} = E_p(O_{i})\, \forall \, O_{i}  \in Z_t$.
We use a learned positional embedding in PyTorch~\cite{NEURIPS2019_9015} to encode the segment identifier as $I_{i} = \texttt{Emb}_{p}(i)$.
We then concatenate the feature vector with the positional embedding identifier for each object,
$P^{\prime}_{i} = P_{i} \oplus I_{i}$.
To improve the generalization ability we randomly generate the object IDs during training~\cite{cui2022positional} over a range larger than the highest number of objects expected to be seen at deployment.
The network then passes these updated features through an encoder $E$ generating a latent feature $\glatentseg_{i} = E(P^{\prime}_{i})$ for each segment. The combined output of all encoders forms the latent state \(\glatent_t\).%

\tpar{Decoder:} Based on these latent features, we can use our decoder, $D$, to generate all outputs associated with the current time step.
We have two distinct kinds of decoders (1) relational decoders,  $D_{r}$, and (2) an environment identity decoder, $D_{e}$.
The relational decoder predicts all segment-segment relations using a set of binary relational classifiers $\hat{\rv}_t = D_{r}(\glatent_{t})$.
The environment identity decoder predicts if a given segment is a movable object or an immobile part of the environment, $\hat{\yv}_{t} = D_e(\glatent_{t})$, \(\hat{\yv}\) defines a vector of outputs for all segments.

\tpar{Actions and Dynamics:} We can learn a dynamics function $\delta$ to predict the resulting latent state based on the current latent state and a selected action $\hat{\glatent}_{t+1} = \delta(\glatent_{t}, A_{t})$, where $A_{t}$ contains the discrete skill to use and its associated action parameters, \(\theta\). We use a discrete parameter to define which object the action will operate on and we encode this into the neural network using the learned positional embedding $\texttt{Emb}_{p}(i)$ for segment \(i\).
We use an action encoder, $E_{a}$, to transform the raw continuous action parameters, \(\theta_c\) sent to the robot controller or motion planner into learned action features, \(\theta_c^\prime\). We denote the total encoded action parameters as \(\theta^\prime = \texttt{Emb}_{p}(i) \oplus \theta_c^\prime\).
We learn a separate dynamics function for each robot skill primitive. This removes the burden of the network having to learn to map skill codes to distinct dynamics outcomes. When needing to be explicit, we will denote the skill specific dynamics as \(\delta_l\) for the dynamics function associated with skill primitive \(A_l\).

\tpar{Latent Space Planning:} Based on the latent state predicted using our learned dynamics function, we can use our decoder to predict the relations at the resulting state \(\hat{\rv}^{\prime}_{t+1} = D_{e}(\hat{\glatent}_{t+1})\). By recursively calling this dynamics function with a sequence of actions $\tau = (A_{0}, \ldots, A_{H-1})$, we can generate rollouts with time horizon \(H\) for use in a planning algorithm to compare the predicted relations with the goal relations.

\subsection{Learning Relational Dynamics with GNNs}
We now turn our attention to our relational dynamics graph neural network, \rdgnn{}, which takes as input the segmented object point cloud and a candidate action and predicts the current and post-manipulation inter-object relations.

Given the output of our point cloud encoder, we define our input graph $\ginput = (V^{I}, E^{I})$ with nodes \(V^{I} = \{E_p(O_i) \oplus k \}\) where \(k\) denotes a one-hot encoding providing a unique identity label for each node.
To improve the generalization ability we randomly generate the object IDs during training~\cite{cui2022positional} over a range larger than the highest number of objects expected to be seen at deployment.

We define edges to and from all node pairs in the graph creating a fully-connected, directed input graph. We set all input edge feature \(\ev_{\ij} \in E^{I}\) to be empty. This topology enables message passing between all nodes, but provides no explicit edge features as input for learning.

We use a graph neural network as the encoder to transform our input graph, \(\ginput\) to a latent state $\glatent = E(\ginput)$. Here $E$ represents a layer of graph message passing and aggregation as defined in the previous section.
We use our latent graph embedding as input to three sub-networks: our relational classifier, \(\rv = D_r(\glatent)\), environment identity classifier, \(\yv = D_e(\glatent)\), and our latent graph dynamics function \(\glatentprime = \delta(\glatent, A)\).

We construct our relational classifier as a  multi-layer perceptron (MLP) that operates on a pair of nodes and their associated edges from \(\glatent\), taking the form of an edge aggregation network \(\rv_{\ij} =  D_{r}(\vv_i, \vv_j, \yv_{i}, \yv_{j}, \ev_{\ij})\).
We predict relations for all object pairs by running this classifier for each pair of nodes in the graph as a form of graph convolution. While some relations may be mutually exclusive, in general the spatial relations are independent of one another, necessitating individual binary classifiers and not a softmax-based multi-class classifier. Note we never specify mutually exclusive goal relations.
We use another MLP to predict environment identity as \(\yv_{i} =  D_{e}(\vv_i, \yv_{i})\).

We additionally examine learning to predict the object pose (defined as its centroid and bounding box orientation in simulation) for all objects in the scene. To this end we learn a pose regressor \(\xv_i = D_p(\glatent)\) which we train using a node aggregation network with an output MLP with 3 outputs encoding position and 6 encoding orientation as in~\cite{zhou-cvpr2019continuity}.

The final piece to define is our latent graph dynamics function \(\glatentprime = \delta(\glatent, A)\).
Recall that \(A\) defines the action (skill) including its skill parameters being evaluated through the dynamics. We encode any discrete skill variables (e.g. object identity) using a one-hot-encoding for use as input into the network.
We pass this action through an action encoder \(A^{\prime} = \actionenc(A)\) which we implement as an MLP.
We build separate node \(\delta_{v}(\cdot)\) and edge \(\delta_{e}(\cdot)\) dynamics functions which respectively take as input the node or edge features of the latent graph concatenated with the encoded action. As output they predict the change in graph features \(\Delta \vv_i^{L}, \Delta \ev_{\ij}^{L}\). Given these definitions we define our graph dynamics functions as
\(\vv_i^{L^\prime} = \vv_i^{L} + \delta_{v}(\vv_i^{L} \oplus \actionenc(A))\)
and
\(\ev_{\ij}^{L^\prime} = \ev_{\ij}^{L} + \delta_{e}(\ev_{\ij}^{L} \oplus \actionenc(A))\).
We incorporate multiple skills by learning a separate dynamics functions for each skill, using the same shared latent space.

\begin{figure*}[ht]
    \centering
    \includegraphics[width=1.98\columnwidth]{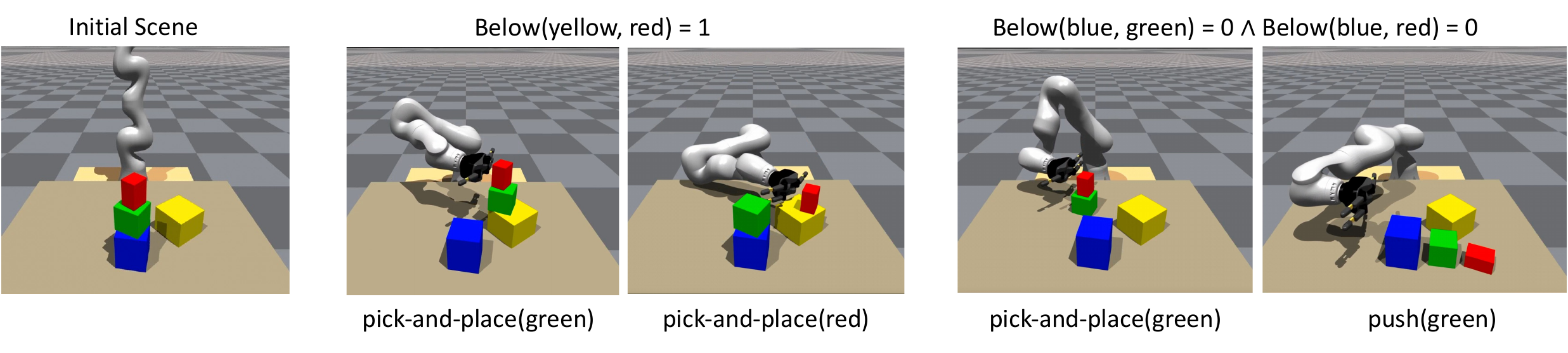}
    \caption{For the same initial scene (left) we show different valid states found by our planner and model for two different goal settings.
    For the first goal relation, the robot can either pick the green object or the red object to place atop the yellow object.
    For the second goal relation, the robot can either push the green object or pick-and-place the green object to deconstruct the towers.
      }
    \vspace{-18pt}
    \label{fig:plan_distribution}
\end{figure*}

\subsection{Transformers for Environment-Object Relational Dynamics}
Based on the high-level introduction of our approach in section~\ref{high_level_approach}, we  define our novel \emph{environment-aware} relational dynamics transformer as \etedyn{}. 
Following the general structure outlined above we use a transformer for the encoder and dynamics, but use MLPs for the decoder. 
This matches the structure of the previously proposed RD-GNN, but replaces the primary GNN components with a transformer. 
The encoder reads in the processed object segments and learned IDs \(P^\prime\) in an arbitrary order with no causal masking.  We concatenate all of the \(N\) latent state feature vectors together to give the full latent state \(\glatent_t = \glatentseg_1 \oplus \glatentseg_2 \oplus \ldots \oplus \glatentseg_N\).
For encoding the input to the transformer dynamics model, we append the encoded action to the end of the sequence of \(N\) latent segment codes and output a sequence of length \(N\) to predicted the subsequent latent state.
For decoding, we follow the recent literature~\cite{yuan2022sornet} to use a simple MLP to read out relations and environment identity predictions associated with each segment.

 \subsection{Loss Functions for Training}
We train our model end-to-end using a combination of loss function terms.
 The first set of loss functions are quite standard aiming to predict the desired labels for each scene observation encoded through the latent space.
 Assuming our training dataset has \(K\) samples, the first loss term is our additional cross entropy loss between predicted environment identity and ground truth identity \(\mathcal{L}_{\text{E}} = \sum_{k=1}^{K}\texttt{CE}(y_{k}, D_{e}(E(E_{p}(Z_k))))\).
 For the current sample we evaluate an additional a cross-entropy loss between predicted and ground truth relations
 \(\mathcal{L}_{\text{R}} = \sum_{k=1}^{K}\texttt{CE}(r_{k}, D_{r}(E(E_{p}(Z_k))))\).
 For pose estimation we define an L2 loss on the current and predicted object poses in an analogous manner replacing $D_r$ with $D_p$.
We examine the effect of this loss in our experiments.
Note the environment identity loss only applies to environment-aware models. 

The second form of loss function acts to ensure agreement in the latent space between states encoded from perception and those predicted through the learned latent dynamics function.

For each observed point cloud $Z_t$, we have a latent state encoding of $\glatent_t = E(E_{p}(Z_t))$. Since datasets are collected in several sequences, we can observe the effects of applying an action sequence \(\tau_t\) of length \(H\)
by computing the output of the dynamics function \(H\) times recursively. We denote the resulting latent space sequence as $\glatentprime_{t+H} = \delta(...,\delta(\glatent_t, {A_{1}}),..., A_{H})$, where we use the ``prime'' symbol to denote prediction though the dynamics function.
We can then define our latent space loss as the L2 norm between $\glatent_{t+H} = E(E_{p}(Z_{t+H}))$ and $\glatentprime_{t+H}$ as \(\mathcal{L}_{\text{d}} = \sum_{i=t}^{t+H-1} \sum_{j=i}^{H}||\glatent_{i+j} - \glatentprime_{i+j}||_2^2\). Here the outer sum ensures we compute the loss with a latent space encoding starting for each observed scene in the sequence.

Based on each predicted latent state $\glatentprime_{t+j}$, we can also decode from the latent state to our supervisory signals of  environment identity and relations using the cross-entropy loss.
We train our model end to end using the sum of all of these loss terms with the Adam optimizer.

\section{Graph search for Efficient Multi-Step Planning}
Our planning problem requires the robot to find a sequence of discrete skills with associated skill parameters, \(\tau=((A_o, \theta_0),\ldots, (A_H, \theta_H))\) that lead from the observed initial scene \(Z_0\) to a state that satisfies a given conjunction of goal relations $\gv = r_{1} \land r_{2} \land ... \land r_{M}, r_j \in \mathcal{R}$. We encode the observed scene into an associated latent state using the learned encoder \(\xv_0 = E(Z_0)\) and use the learned latent dynamics model to predict the latent states resulting from selected actions. We can then predict the relations in the resulting latent state using the learned decoder and compare them to the goal to evaluate if the plan succeeds.
This results in a mixed-integer programming problem, where the discrete choices of which skill, \(A_l\), to use selects a different associated continuous dynamics function, \(\delta_l(\cdot)\). The RD-GNN avoided this complexity by assuming access to a plan skeleton of logical subgoals, $G = (\bar{\gv}_{1}, \ldots, \bar{\gv}_{H})$, that if achieved would result in satisfying all goal relations. We note that this access to a skeleton has been a common first step in researching complex learning-based manipulation tasks~\cite{lozano2014constraint, simeonov2020long}.

\begin{figure}[h!]
  \centering
  \includegraphics[width=0.8\columnwidth]{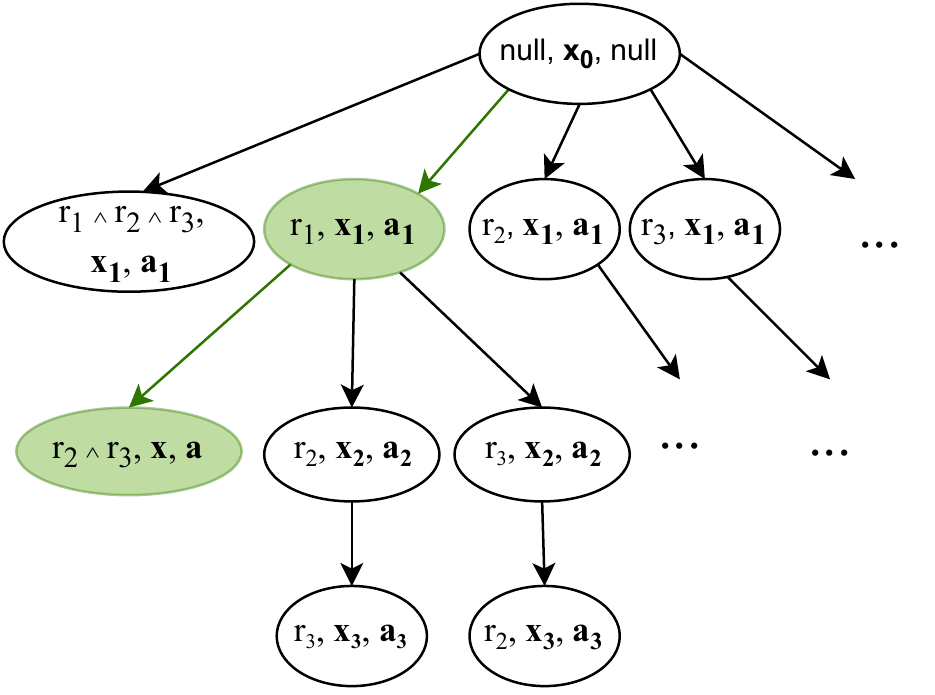}  %
  \caption{Visualization of our logical subgoal graph search. The root node of the tree contains the initial state encoded from the observed scene as well as an empty subgoal and null action. The search prioritizes longer subgoals first to induce shorter plans. The green shaded nodes represent satisfied subgoals. If a satisfied subgoal matches the given goal the search ends.}\label{fig:graph_search}
\end{figure}

To avoid the need for this plan skeleton, we improve on the approach of~\cite{Huang-icra2023-graph-relations} by proposing a graph search to enumerate combinations of goal relations to create plan skeletons. 
We note that given a state, \(\xv_t\) and associated subgoal \(\bar{\gv}_{t+1}\) we can solve a simpler one-step optimization problem to find an action, \(A_t\) and associated parameters \(\theta_t\) that our learned model predicts most likely satisfy the relations in the subgoal. We can then perform a graph search over subgoals where we store in each node the subgoal, as well as the one step action and resulting latent state that satisfy the subgoal. If the one-step optimization fails to find an action that achieves the subgoal with high confidence, we mark the subgoal as visited and remove it from consideration. If the single-step planning problem results in a high confidence prediction of relations that satisfy the goal, we continuously search the remaining goal relations. After all goal relations $\gv$ are satisfied, the search ends.
We use an iterative-deepening style search~\cite{korf1985depth} to bias towards shorter subgoal sequences. 
We visualize our graph search procedure in Fig.~\ref{fig:graph_search}.

To solve our single-step planning problem we use the cross-entropy method (CEM) planning approach. This uses the derivative-free, sampling-based CEM solver to iteratively evaluate a random set of continuous skill parameters and then fit a new distribution to the top-K samples at each iteration. This distribution is then used to sample actions for the subsequent iteration. We found that we only needed 2 iterations of CEM to find good skill parameters if the correct discrete skill and object of interest were selected. The solver runs independently for each possible discrete skill and object pair.
Note since we have learned using our environment identity classifier whether a given segment is manipulable or immobile, we can prune from consideration segments predicted as immobile. We perform experiments in Section~\ref{sec:experiments} to quantify the advantage of this approach.

\section{Experiments with Inter-Object Interactions}
\begin{figure*}
    \centering
    \includegraphics[width=0.66\columnwidth,clip,trim=3mm 0mm 10mm 6mm]{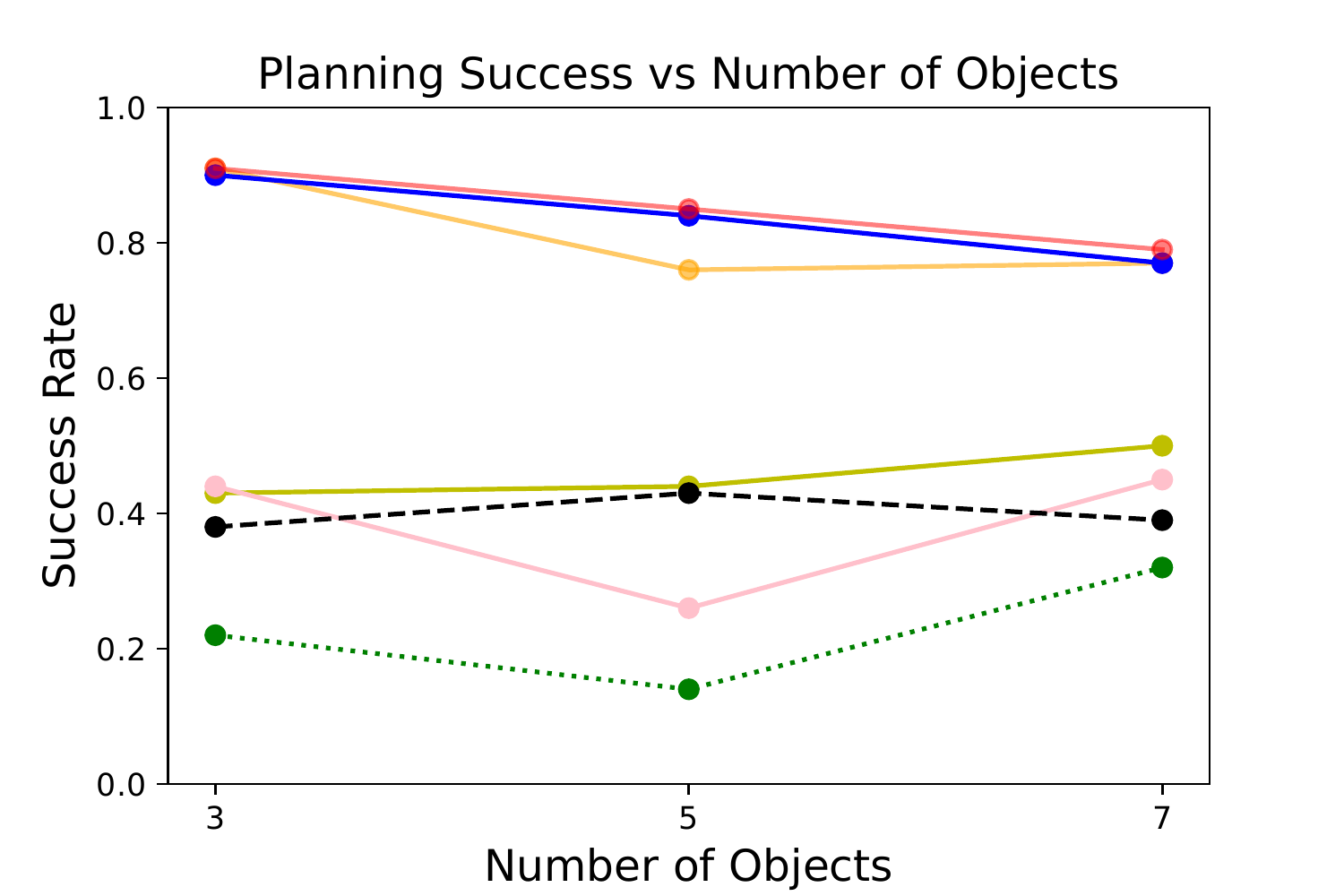} %
    \includegraphics[width=0.66\columnwidth,clip,trim=3mm 0mm 10mm 6mm]{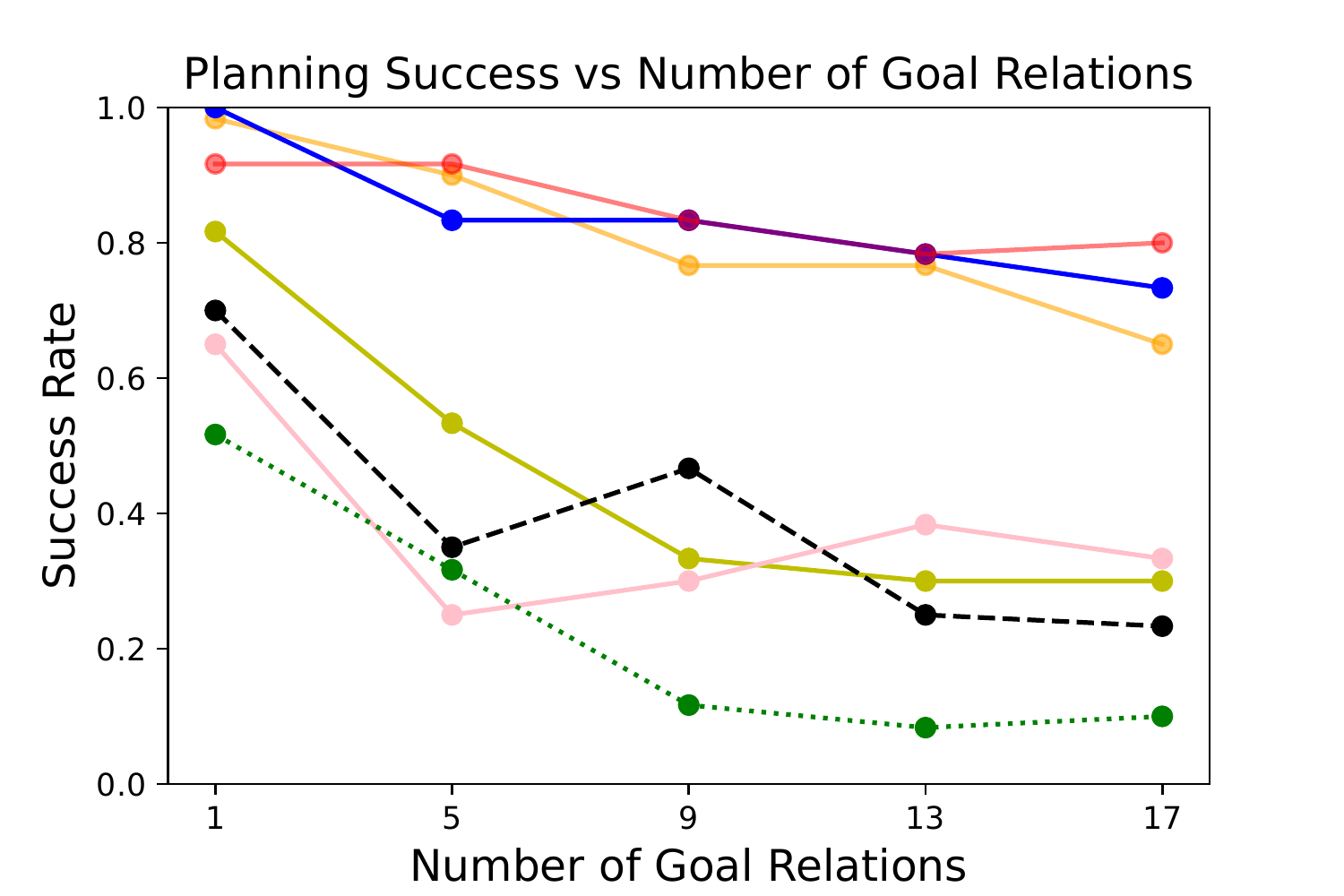} %
    \includegraphics[width=0.66\columnwidth,clip,trim=3mm 0mm 10mm 6mm]{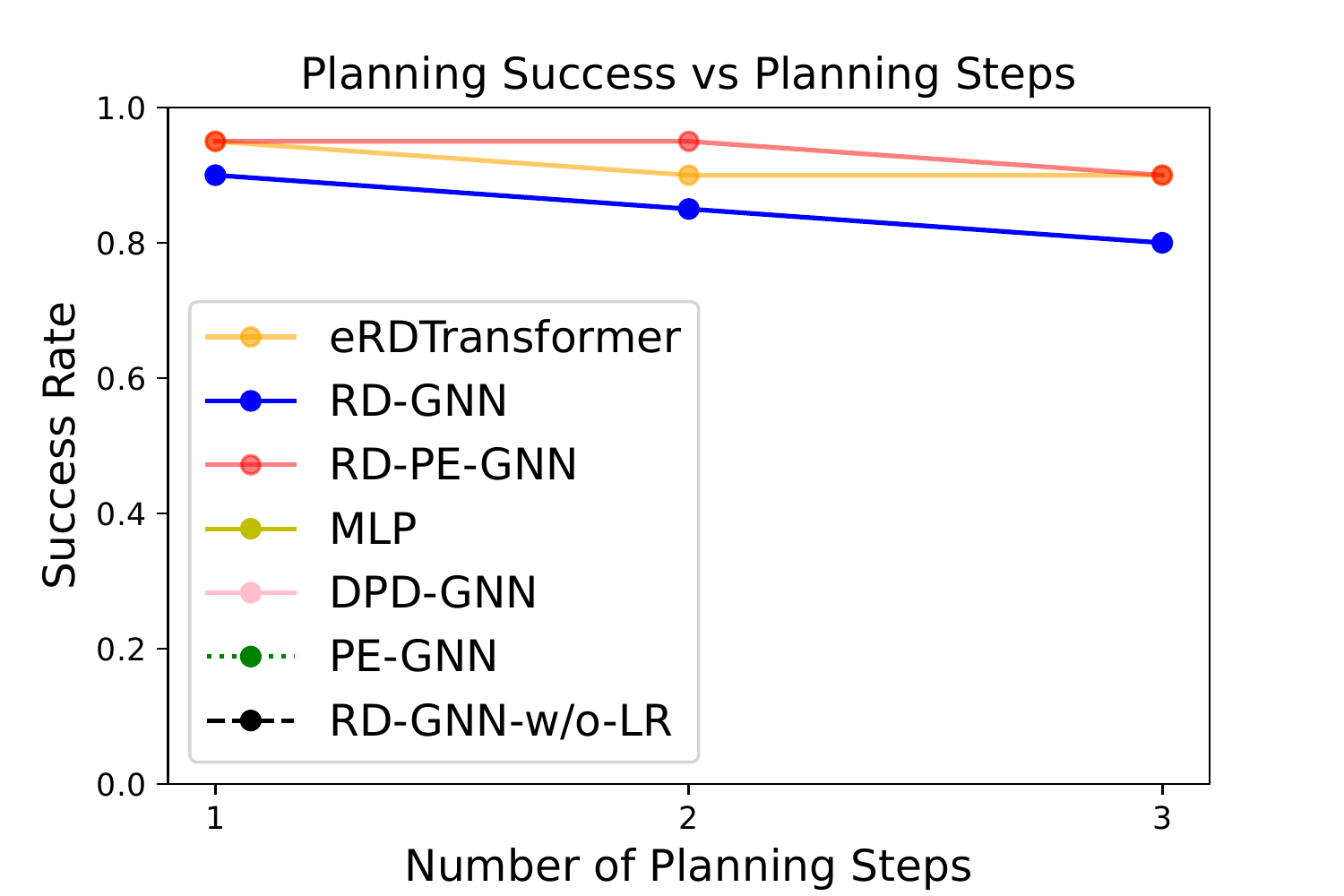}
    \caption{Comparing planning success rate of the different models as a function of (left) the number of objects in the scene, (middle) the number of relations specified in the goal, and (right) the number of steps. The legend applies to all three plots. We see that \etedyn{}, RD-GNN, and RD-PE-GNN achieve comparable performance while significantly outperforming the baseline models. The success rate drops for all models as we specify more relations in the goal. Even when fully constrained the top performing models achieve high success rates.}\vspace{-2pt}
    \label{fig:planning_success_comparison_no_env}
\end{figure*}

We now describe the training data collection and experimental validation for our approach to learning and planning with \etedyn{} and \rdgnn{}.
In our experiments we examine the following relations: \texttt{left}, \texttt{right}, \texttt{behind}, \texttt{in-front}, \texttt{above}, \texttt{below} and \texttt{contact}.
We get the \texttt{contact} relation directly from simulation and we define other relations following Paxton et al.~\cite{paxton-corl2021-semantic-placement}.
We train and evaluate our model on multi-object rearrangement tasks using pushing and pick-and-place skills.
We conduct experiments in simulation and on a physical robot manipulating both blocks and YCB objects~\cite{calli2015ycb}.

\subsection{Dataset Collection}
We conduct large scale data collection using the Isaac Gym simulator~\cite{isaacgym}.
We collect a dataset by generating scenes with a variable number of cuboid objects of random size with arbitrary pose. 
In the training dataset, we have a maximum of 6 objects on the table.
We then execute a random push or pick-and-place skill on one of the objects in the scene. 
For the push skill, the robot randomly pushes one object in the scene with random direction and length. 
For the pick-and-place skill, the robot randomly picks one object in the scene and places the object in a random pose. 
We record the partial view point cloud before and after the manipulation, the executed action, and the ground truth relations between all object pairs in the scene. 
We collected a total of 39,600 push and pick-and-place attempts.
Fig.~\ref{fig:plan_distribution} shows an example scene with various pushing and pick-and-place actions and outcomes from the simulator.

\subsection{Baseline Approaches} We implement several baselines for comparison to our proposed models \emph{\etedyn{}} and \emph{\rdgnn{}} .
\emph{Pairwise MLP Relational Dynamics (MLP)}: predicts relations and dynamics for pairs of objects using an MLP instead of a GNN or transformer to construct the latent space and dynamics. This baseline acts as an analogous model to that in~\cite{paxton-corl2021-semantic-placement}.
\emph{Direct Pose Dynamics GNN (DPD-GNN)}: uses a GNN to predict the pose for each object conditioned on a chosen action. We use an analytic relational classifier to predict relations from the predicted poses and their associated bounding boxes. This acts as a graph-neural-network model similar to the point-cloud dynamics model from~\cite{simeonov2020long}.
\emph{Pose Estimation GNN (PE-GNN)}: We replace the relational output heads on our model with pose estimation regressors. We again use analytic relational classifiers for evaluation.
\emph{Combined Relational Dynamics and Pose Estimation (RD-PE-GNN)}: This combines our model with the pose estimation regressor for both the current and next time step.
\emph{Relational Dynamics without Latent Regularization (RD-GNN-w/o-LR)}: We train a version of our model without using the latent space loss $\mathcal{L}_{\text{d}}$.

\subsection{Evaluation Metrics} We first examine the efficacy of our model in correctly predicting which relations will be present after executing a specified action. 
We show the prediction accuracy between the predicted relations and the ground truth relations post-manipulation.
Then we examine the ability of our model to detect inter-object relations for objects in the observed scene post-manipulation. 
We show the F1 score between the detected and ground truth relations post manipulation. 
In the experiments, we show the average prediction F1 score and detection F1 score across all relational classifiers predictions on our simulation test data across 300 skill executions. 
We use the F1 score, rather than accuracy, as our evaluation metric, because there are imbalanced class distributions in terms of true and false relations in the dataset.

Finally, we examine the ability of our model to desired goal relations with a single action step or multiple action steps. 
We ran 20 planning trials containing varying numbers of objects and goal relations using each model in simulation.

\begin{table}[h!]
\centering
\caption{F1 score for inter-object relational dynamics evaluation}
\begin{tabular}{ p{3cm}p{1.8cm}p{1.5cm}p{1.5cm}}
 \toprule
 Approaches & Prediction & Detection\\
 \midrule
 eRDTransformer   & 0.899  & 0.961 \\
 RD-GNN  & 0.906 & 0.974 \\
 RD-PE-GNN  & 0.879 & 0.977 \\
 PE-GNN  & 0.133 & 0.899 \\
 DPD-GNN  & 0.319 & N/A \\
 MLP  & 0.678 & 0.971 \\
 RD-GNN-w/o-LR  & 0.693 & 0.977 \\
 \bottomrule
\end{tabular}
\label{table:evaluation_f1_no_env}
\end{table}

\subsection{Inductive Bias of Transformers and GNNs}
We first examine whether the transformers and GNNs have better inductive bias compared to the MLP baseline. 
The average prediction and detection F1 score for \etedyn{}, \rdgnn{}, and MLP is shown in table~\ref{table:evaluation_f1_no_env}. 
We found that \etedyn{} performs comparably to \rdgnn{} while outperforming MLP baselines in terms of prediction F1 score.
All three approaches perform great with average detection F1 score. 
In terms of planning success rate in Figire~\ref{fig:planning_success_comparison_no_env}, we find again \etedyn{} and \rdgnn{} perform much better than the MLP baseline. 
Fig.~\ref{fig:plan_distribution} shows a variety of successfully executed single-step plans using RD-GNN. Notably we generate diverse plans for the same goal and initial setting.

This evaluation shows the effectiveness of using transformers and GNNs to address multi-object relational dynamics. We attribute the better results to the inductive bias of the transformers and GNNs.

\subsection{Effects of Relational Supervision and 
Latent Regularization}
In this experiment, we aim to understand whether relational supervision and latent regularization are important. 
We compare our model \rdgnn{} to several baselines using pose supervision DPD-GNN, PE-GNN, and RD-PE-GNN.  
Additionally, we compared our model to a variant without latent regularization RD-GNN-w/o-LR. 
We show the average prediction F1 score and detection F1 score in table~\ref{table:evaluation_f1_no_env} while showing the planning success rate in Fig.~\ref{fig:planning_success_comparison_no_env}. 
During the comparisons, we found \rdgnn{} performs better than PE-GNN, DPD-GNN, and RD-GNN-w/o-LR while performing comparably to RD-PE-GNN in terms of prediction F1 score and planning success rate. 
All the approaches perform pretty well in terms of detection F1 score except PE-GNN. 

The comparisons between \rdgnn{}, RD-PE-GNN, PE-GNN, and DPD-GNN demonstrated that using relational supervision is better than using pose supervision and training with both the relational and pose estimation loss provides no real benefit over training with only relational loss. 
Furthermore, The comparison between \rdgnn{} and RD-GNN-w/o-LR shows the importance of latent regularization loss during model training.

\begin{figure}[ht]
  \centering
  \vspace{-5pt}
  \includegraphics[width=0.98\columnwidth,clip,trim=15mm 0mm 10mm 8mm]{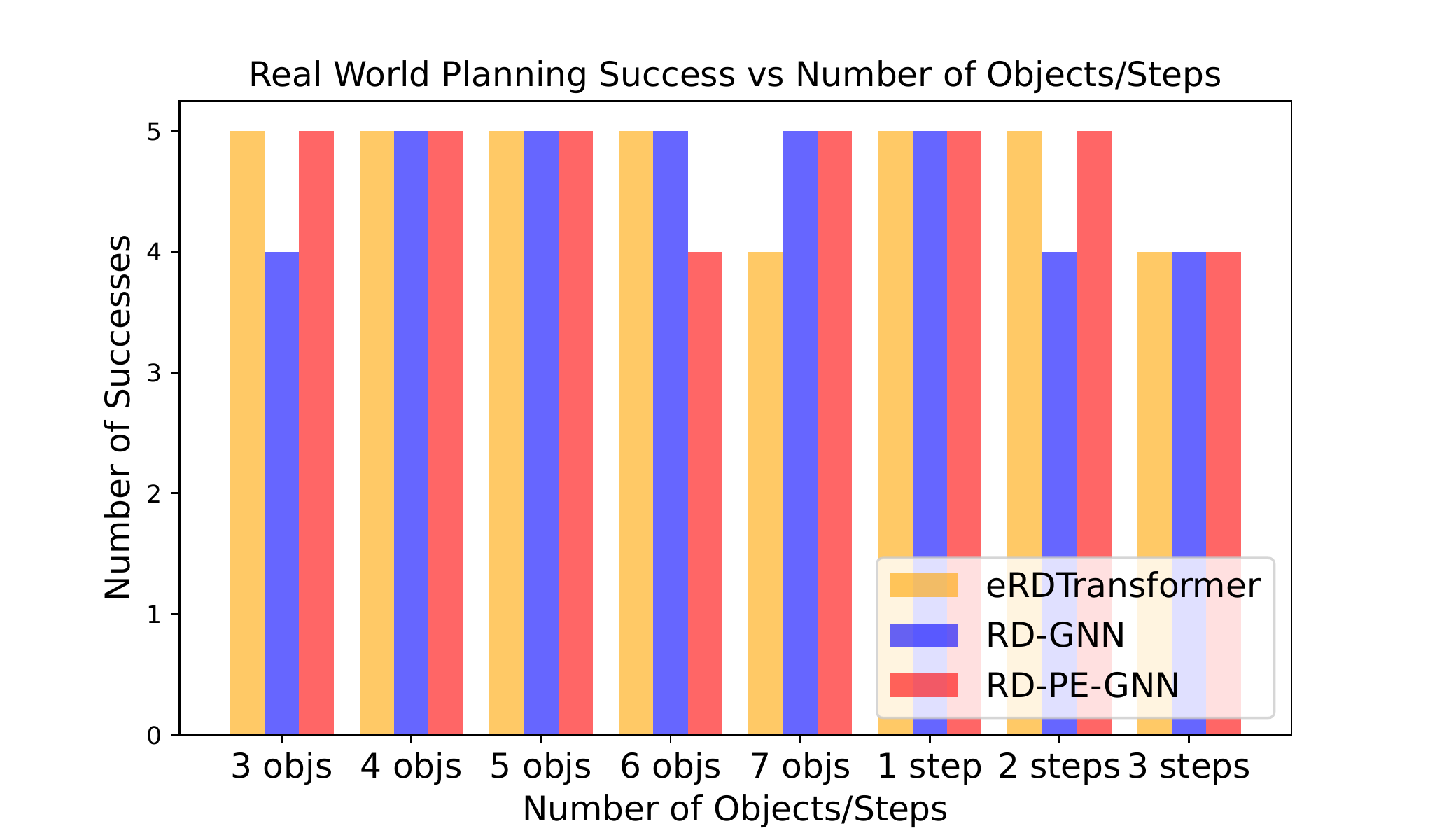}  %
  \caption{Number of successes on real world YCB object manipulation tasks. We compare results for a varying number of objects and varying plan horizon length as denoted by the horizontal labels. We ran 5 trials for each task per model.}\vspace{-0pt} \label{fig:real_all_no_env}
\end{figure}

\subsection{Multi-Step Planning and Real-World Experiments}
We now turn our attention to multi-step planning and real world planning focusing on only \etedyn{}, RD-GNN, and RD-PE-GNN as the best single step performers. We show planning success rates for plans ranging in length from 1 to 3 planning steps in simulation in Fig.~\ref{fig:planning_success_comparison_no_env} (right).
We use four objects for the multi-step test in simulation and the real world.
We see that for both \etedyn{}, \rdgnn{}, and RD-PE-GNN approaches, the success rate drops with plan length. All models achieve high success rates.

We show planning success rate for real world experiments in Fig.~\ref{fig:real_all_no_env}.
We ran test trials using YCB objects~\cite{calli2015ycb} with 5 trials for each setting of varying number of scene objects or plan steps.
For all the real world experiments, we use 5 relations in the goal.
Our results verify that our methods transfer to real world settings without any fine-tuning and generalize to real world objects when trained only on cuboids in simulation.

\begin{figure*}[ht]
  \centering
  \vspace{-5pt}
  \includegraphics[width=1.98\columnwidth,clip,trim=0mm 0mm 0mm 0mm]{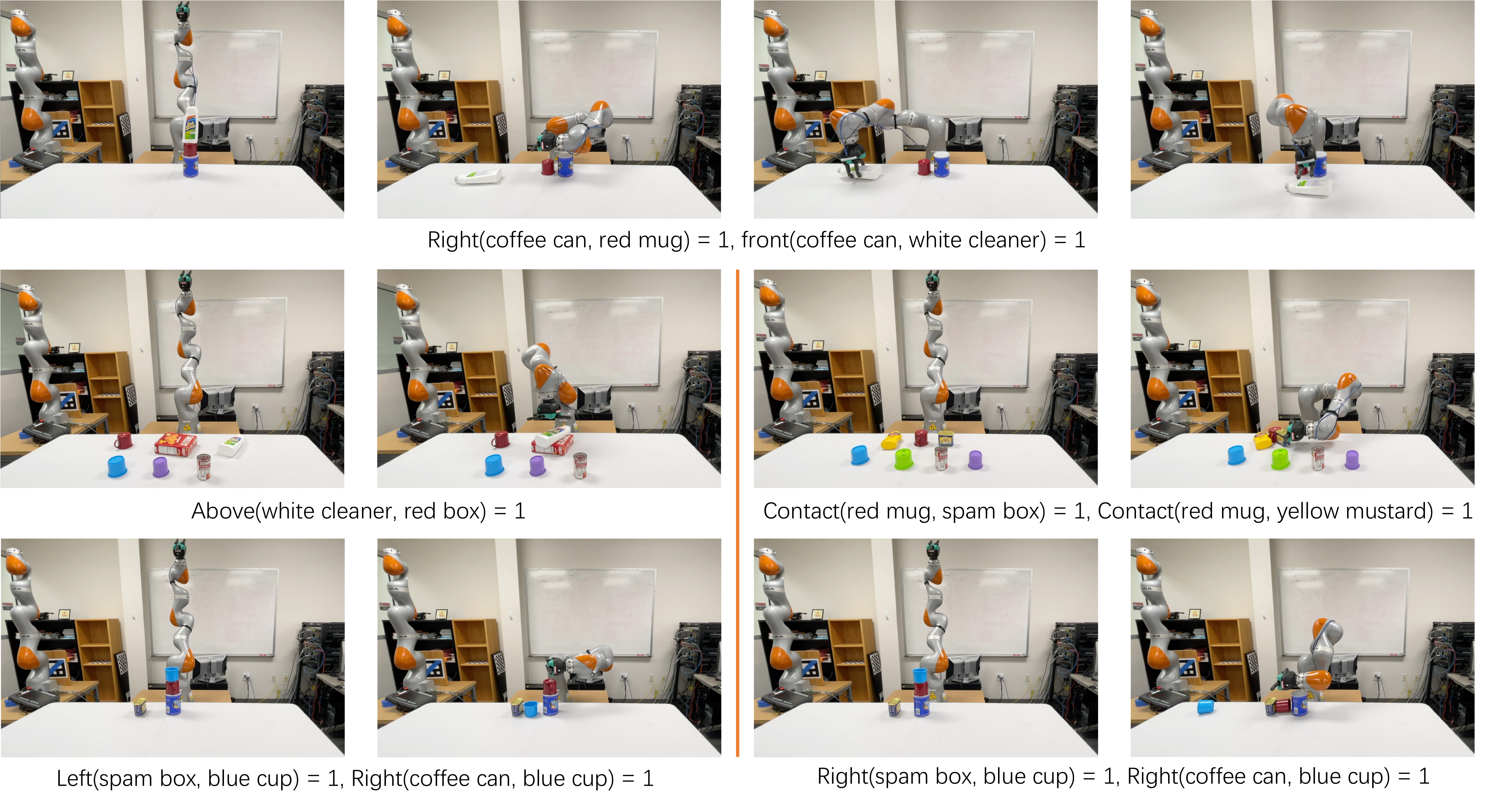}  %
  \caption{We show several multi-step planning executions (top) and single-step planning executions (middle, bottom) in the real world using our approaches.
  We can reason about a different number of objects with varied geometry.  
  Note relations are defined from the camera's perspective.}\vspace{-5pt} \label{fig:visualization_multi_objs}
\end{figure*}

We qualitatively show several real-world plan executions in Fig.~\ref{fig:visualization_multi_objs}. 
These visualizations show that our framework can reason about inter-object interaction and plan to achieve goal relations with different numbers of objects in the real world. 
Even though our training dataset includes a maximum of 6 cuboids, our tests show our approach can generalize to different YCB objects including scenes with more than 6 (shown for 7) objects.
As shown in the bottom row of Fig.~\ref{fig:visualization_multi_objs}, the robot can choose to push either a red mug or a blue cup to achieve different goal relations with the same initial scene. 
These results highlight our framework's ability to reason about different goal relations.

\section{Experiments with Object-Environment Interactions}\label{sec:experiments}

In this section, we show the experiments to address object-environment interactions using our approach \etedyn{}. 
We train and evaluate our approach based on the following ground-truth relations definition:
\texttt{left}, \texttt{right}, \texttt{behind}, \texttt{in-front}, \texttt{above}, \texttt{below}, \texttt{contact}, and \texttt{boundary}.
We get \texttt{contact} directly from physics-based simulation.
We define the \texttt{boundary}(A, B) = 1 if \texttt{above}(A, B) = 1, the center of A to the nearest edge of the bounding box of B is less than 10 cm, and the smallest extent of bounding box of B is larger than 20 cm.
For other relations, we define them the same way as~\cite{paxton-corl2021-semantic-placement}.

\subsection{Dataset Generation}
\begin{figure*}[ht]
\includegraphics[width=1.98\columnwidth,clip,trim=0mm 0mm 0mm 0mm]{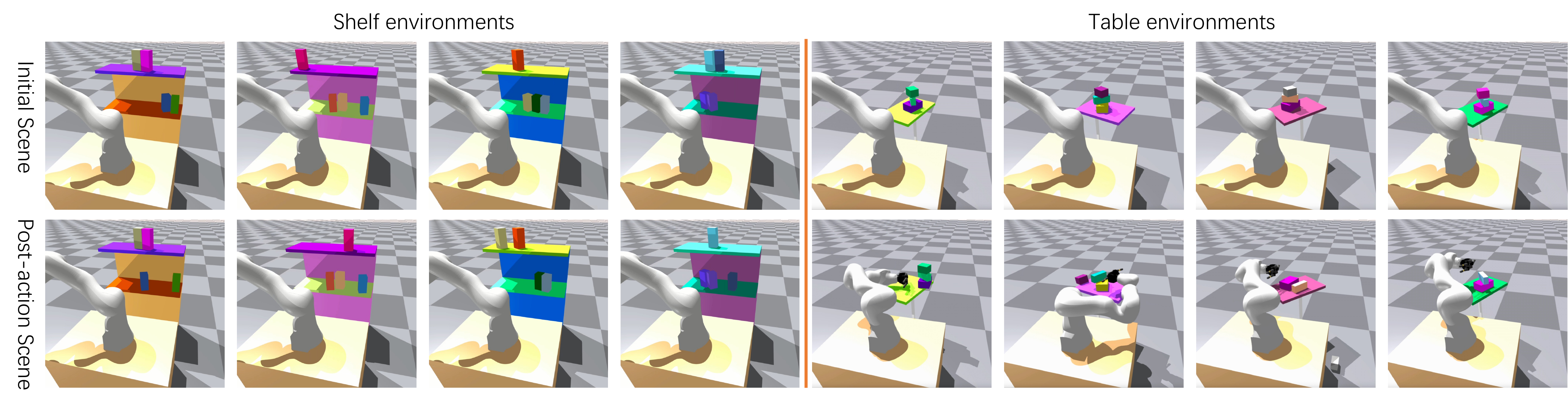}  %
      \caption{Visualization of the simulation dataset collection with table and shelf environments.} \label{fig:visualization_environments}
\end{figure*}

We collect a large training dataset using the Isaac Gym simulator~\cite{isaacgym}.
We randomly generate environments including varied tables and bookshelves containing a variety of cuboid objects of different sizes.
We execute random behavior sequences using both pick-place and push skills.
During the simulated robot executions, we save ground truth segmented partial-view point clouds before and after execution of each robot skill along with the skill executed and the ground truth relations.
We collect in total 13,000 skill executions with different environments.
Fig.~\ref{fig:visualization_environments} visualizes some simulation environments.
For the shelf environments, we have two horizontal shelves and four cuboids on the shelves. The four cuboids have random poses and different sizes. The size and height of the shelves are fixed. Then the robot randomly picks one cuboid and places the cuboid at another pose.
For the table environments, we have three cuboids on top of the table. 
The table and three cuboids have different sizes and poses. 
The three cuboids always make up a stack. Then the robot randomly executes a push skill on one of these three cuboids with random direction and distance.

\begin{figure}[ht]
\includegraphics[width=0.98\columnwidth,clip,trim=0mm 0mm 0mm 0mm]{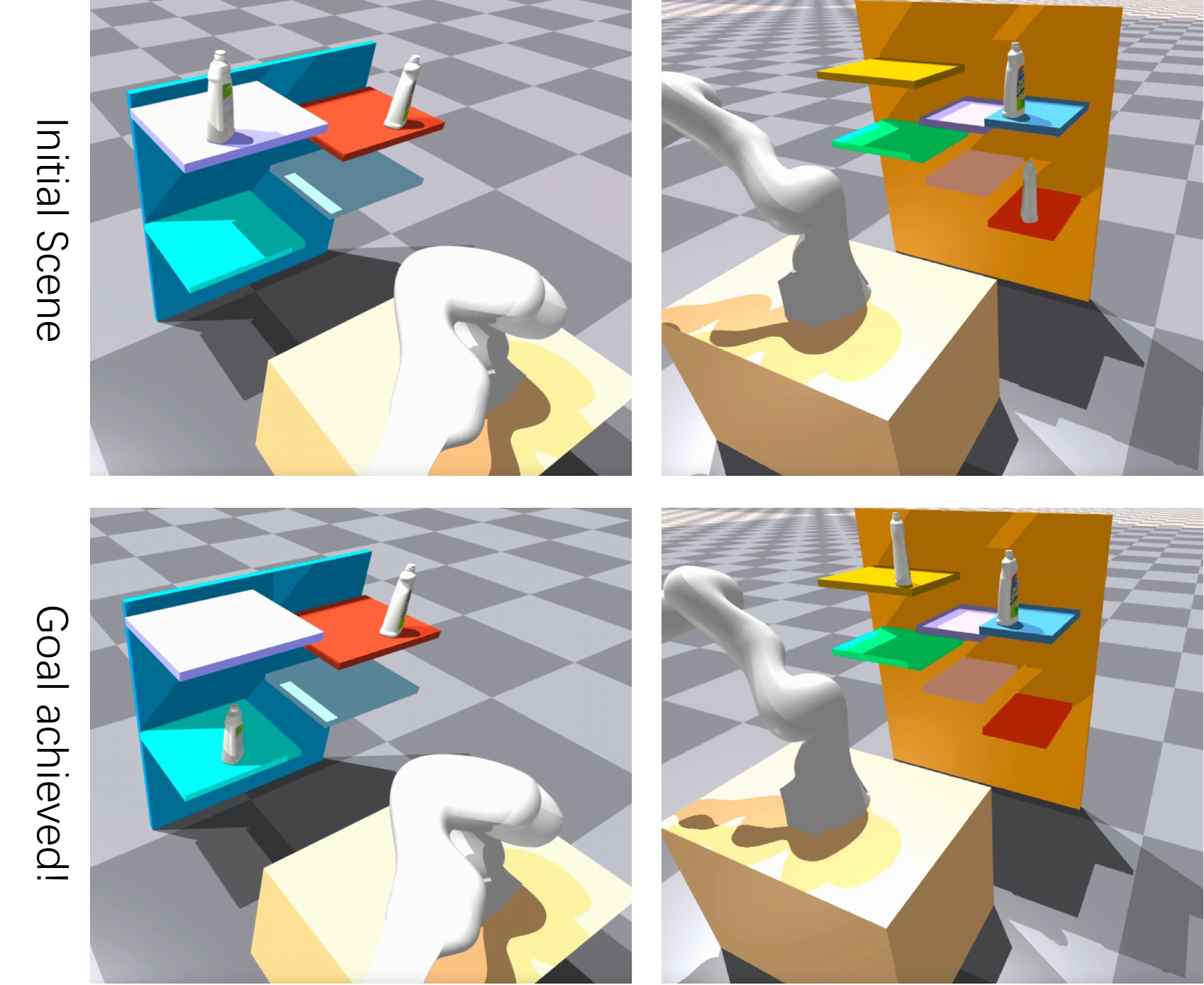}  %
      \caption{Two test examples with more complex shelves and novel view points. For the left column with 4 varied height shelves, the goal is to put the white cleaner that's on the top layer of the shelf to the bottom layer. For the second column with 6 varied height shelves, we would like our robot to put the white cleaner which is on the bottom layer of the shelf to the top layer.} \label{fig:simulation_bookshelf}
\end{figure}

\begin{figure*}[ht]
\includegraphics[width=1.98\columnwidth,clip,trim=0mm 0mm 0mm 0mm]{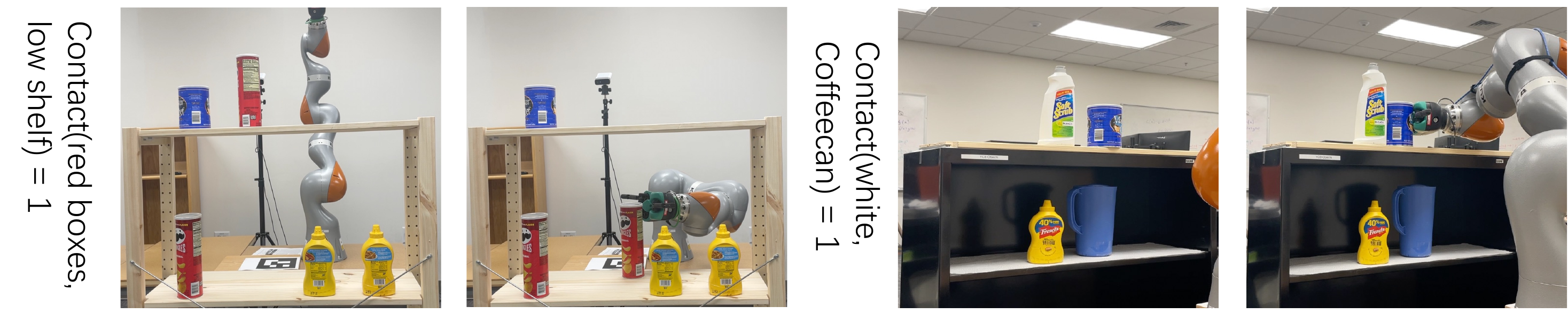}  %
      \caption{Real-world evaluation with different shapes of shelves and different numbers of objects. Our framework achieves goals such as picking and placing an object onto different layers of the shelf and pushing several objects to make contact with one another in a shelf environment.} \label{fig:real_bookshelf}
\end{figure*}

During collection of the dataset objects will sometimes be out of camera view after, e.g. falling off the table.
To get a dense point cloud of objects that are out of the camera view, we augment the observed point cloud at the current step with an object-segment point cloud from the previous timestep rigidly transformed to the resulting object pose.
Note this data augmentation method only applies to training datasets where we need ground truth encoding of all segments at each timestep to compute the latent-space loss term.

\subsection{Basline Approaches}

\tpar{\egnn{}} is the environment-aware extension to the relational-dynamics graph net (\rdgnn{}). The model uses GNN as the encoder and decoder while using MLPs to encode the graph dynamics in latent space. Note in the GNN-based approach, the latent space $\glatent$ is a latent graph that contains latent nodes and latent edges. We implement the same graph net message passing and aggregation functions as in \rdgnn{}.
We implement the relational decoder as an MLP $D_{r}$ to predict the relations between all pairs of nodes in the graph. We further extend over the previous RD-GNN implementation by adding our environment identity decoder, $D_{e}$ to predict the environment identity for each segment.

We define two slightly different models based on \etedyn{} to study the effects on performance with using a transformer for different model components. 

\tpar{\erdt{}} leverages transformers for the encoder, dynamics function as in \etedyn{}. 
But \erdt{} uses an extra transformer in the decoder. 
For decoding with the transformer, we first pass the latent
codes through another transformer decoder and then readout
relations and environment identity predictions associated with
each segment using an MLP. 
This allows us to understand the contribution of the transformer in the decoder.

\tpar{\etemdy{}} serves a similar role to the previous model as an ablation.  
Here we replace the dynamics function from \etedyn{} with an MLP. 
This helps us to understand the importance of using a transformer as the dynamics function.

\subsection{Evaluation Metrics}
We first examine the predictive performance of the proposed models. 
We evaluate both the ability of the models to predict relations and environment identity in the currently observed scene and post-manipulation scene (detection) and the ability to predict how these relations will change (prediction).
Fig.~\ref{fig:visualization_environments} visualizes the environments used for evaluation.
We evaluated these F1 scores on these tasks for all of our models across a varying number of objects with environments containing either one table or two shelves as immobile surfaces.
For each approach, we get our simulation test data with 80 skill executions for evaluation. 

Furthermore, we evaluate how well these models work in the context of manipulation planning for both single-step and multi-steps. 
We ran 20 trials using each model for each environment. 
We used different numbers of goal relations for different environments with a maximum number of goal relations of 17. 

\subsection{Transformer for Object-Environment Relational Dynamics}
Both \etedyn{} and \rdgnn{} have shown great performance in reasoning about multi-object relational dynamics. 
So now we investigate which structure is better for reasoning about object-environment relational dynamics. 
We show the comparisons among \etedyn{}, \rdgnn{}, and \egnn{}. 

Through the comparisons in table~\ref{table:evaluation_f1_env}, we found that these three approaches perform comparably well in terms of env detection F1 score and detection F1 score except \rdgnn{} cannot predict which segment is the environment. 
The main difference comes from the prediction F1 score, the \etedyn{} performs obviously better than \egnn{} and \rdgnn{}. 
In the evaluation of single-step and multi-step planning success rate in Fig.~\ref{fig:planning_success_comparison_env}, we found that \etedyn{} performs better than \egnn{} while \rdgnn{} performs worst. 

From the evaluations, we found that \etedyn{} outperforms \rdgnn{} and \egnn{} in terms of both planning success rates and predicting post-manipulation relations, which demonstrates that the transformers provide better inductive bias than GNNs to explicitly reason about object-environment relational dynamics. 

\subsection{Ablation Study}\label{sec:abalation_study}
\begin{figure}[ht]
    \centering
    \includegraphics[width=\columnwidth,clip,trim=3mm 0mm 10mm 6mm]{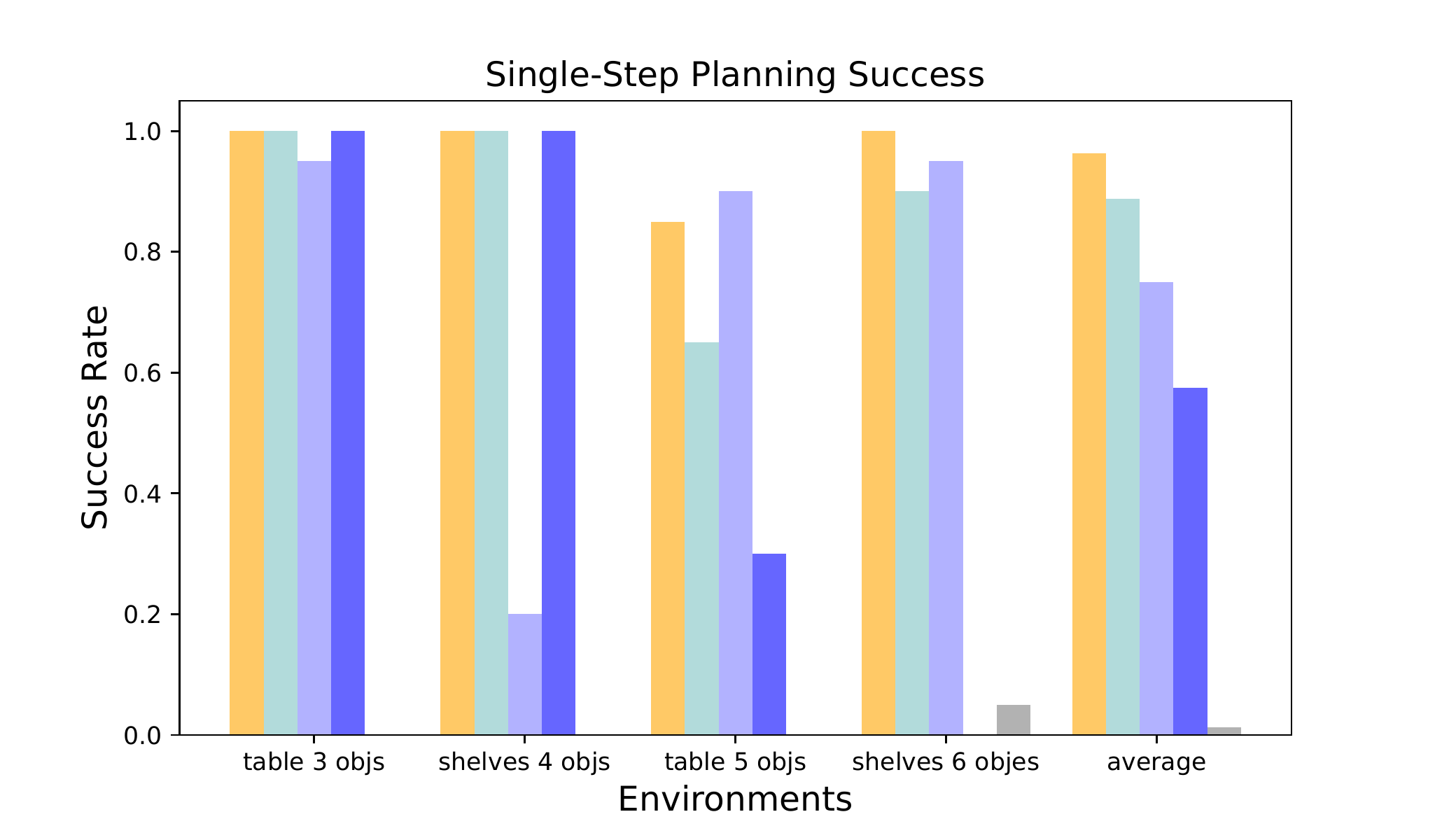} \label{fig:comparison_obj_num}  %
    \includegraphics[width=\columnwidth,clip,trim=3mm 0mm 10mm 6mm]{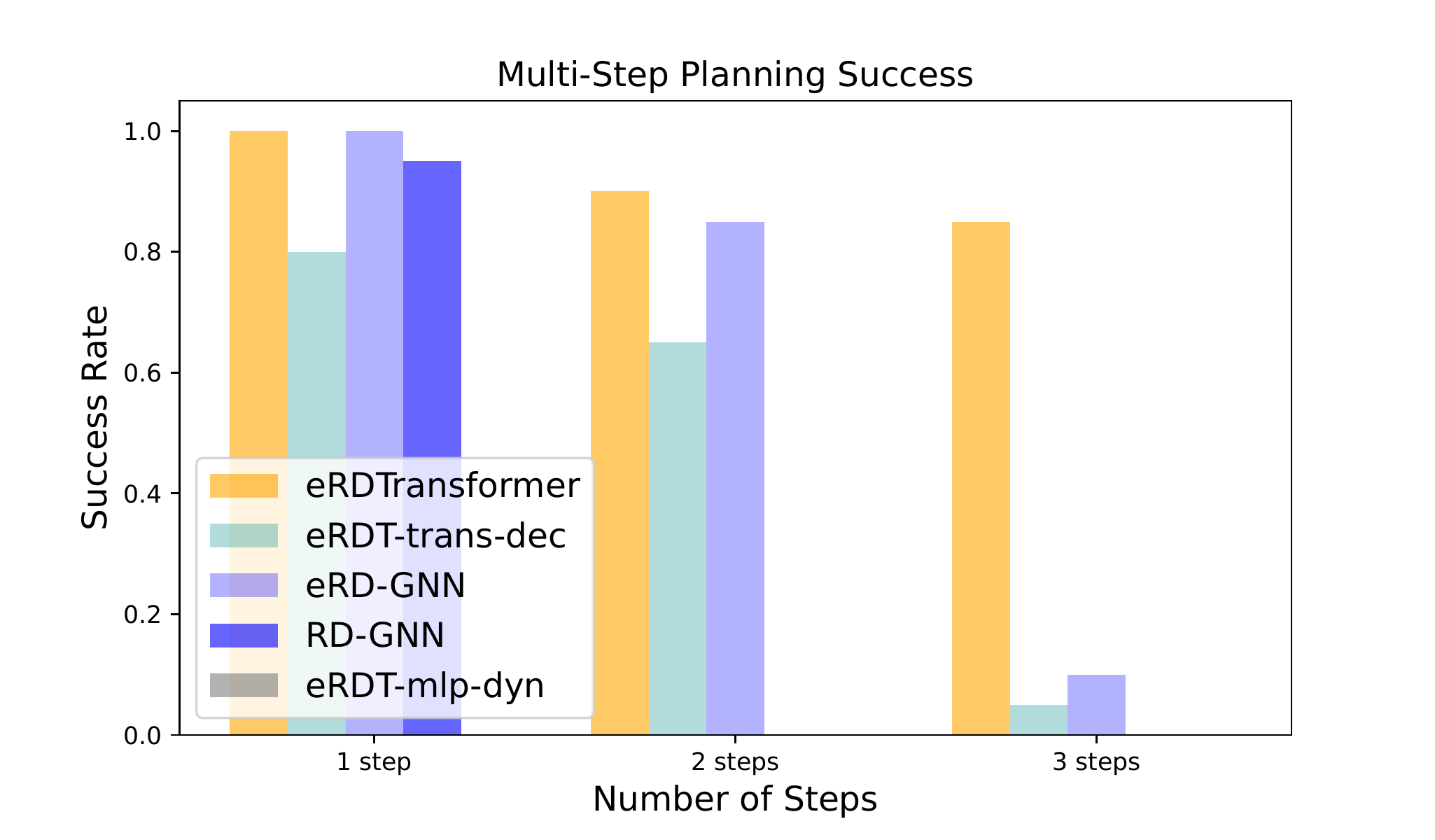}
    \label{fig:multi_step}
    \caption{Execution success rates for single-step (top) and multi-step (bottom) planning in simulation. Empty bars denote 0\% success.}
    \label{fig:planning_success_comparison_env}
\end{figure}

In this experiment, we study the effect on the performance when using a transformer for different model components. 
We show the results to compare our model \etedyn{} with two ablations \erdt{} and \etemdy{}. 
During the comparisons in table~\ref{table:evaluation_f1_env}, we found that all three models perform equally well at env detection and relation detection. 
We found that \etedyn{} performs better than \erdt{} in terms of predicting post-manipulation relations while \etemdy{} performs worst. 
During the evaluation of single-step and multi-step planning success rate in Fig.~\ref{fig:planning_success_comparison_env}, we found \etedyn{} performs best, \erdt{} second, while \etemdy{} performs worst. 

Through the ablation study, we found that using the transformer as the dynamics is most important for our latent space planning framework while using a transformer for the decoder hurts the performance, especially in multi-step planning.

\begin{figure}
    \centering
    \includegraphics[width=\columnwidth,clip,trim=0mm 0mm 0mm 0mm]{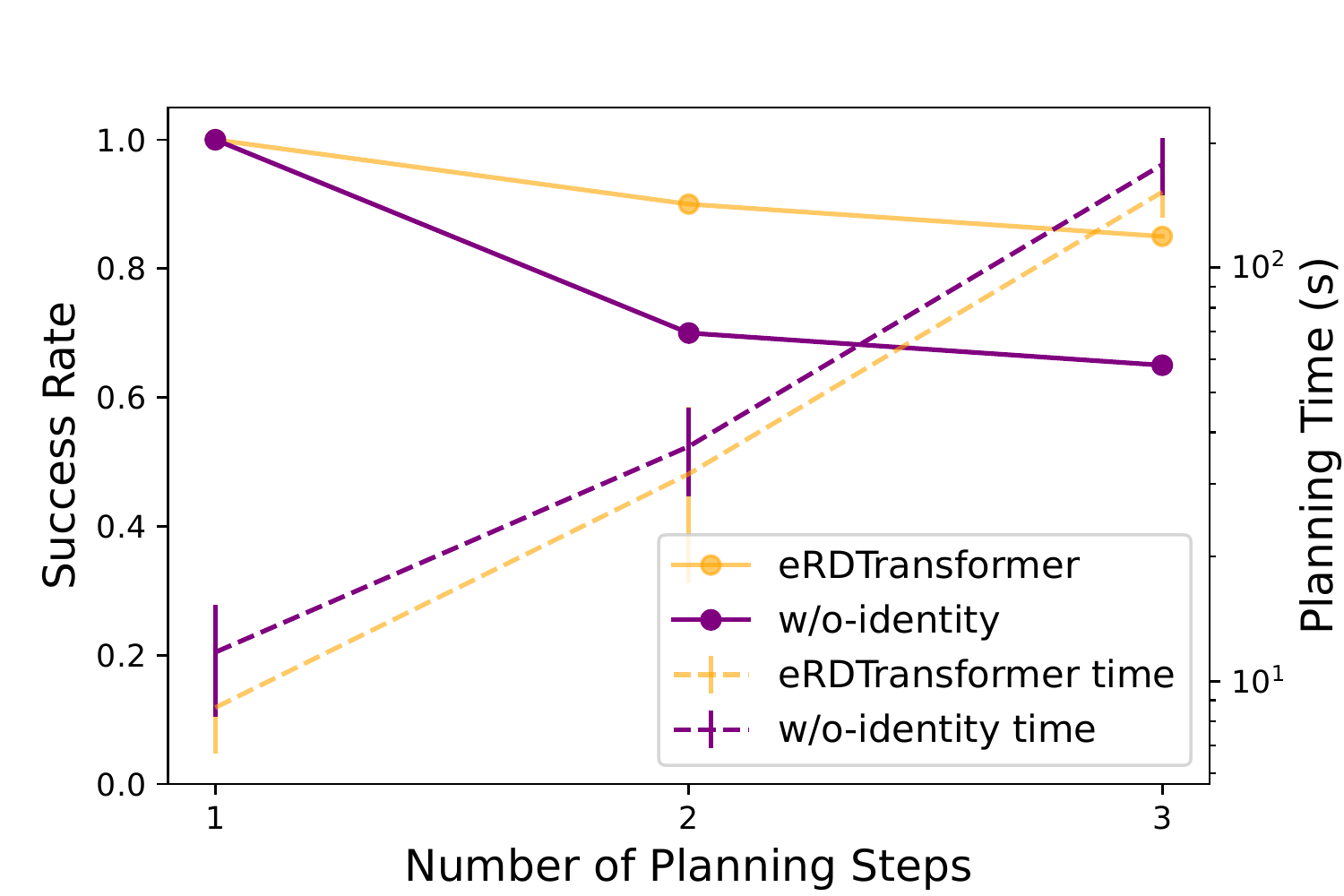}
    \caption{Pruning segments classified as immobile environment regions from search improves planning success and decreases planning time.}
    \label{fig:Env_identity}
\end{figure}

\begin{figure*}[ht]
    \centering
    \includegraphics[width=1.8\columnwidth,clip,trim=0mm 0mm 0mm 0mm]{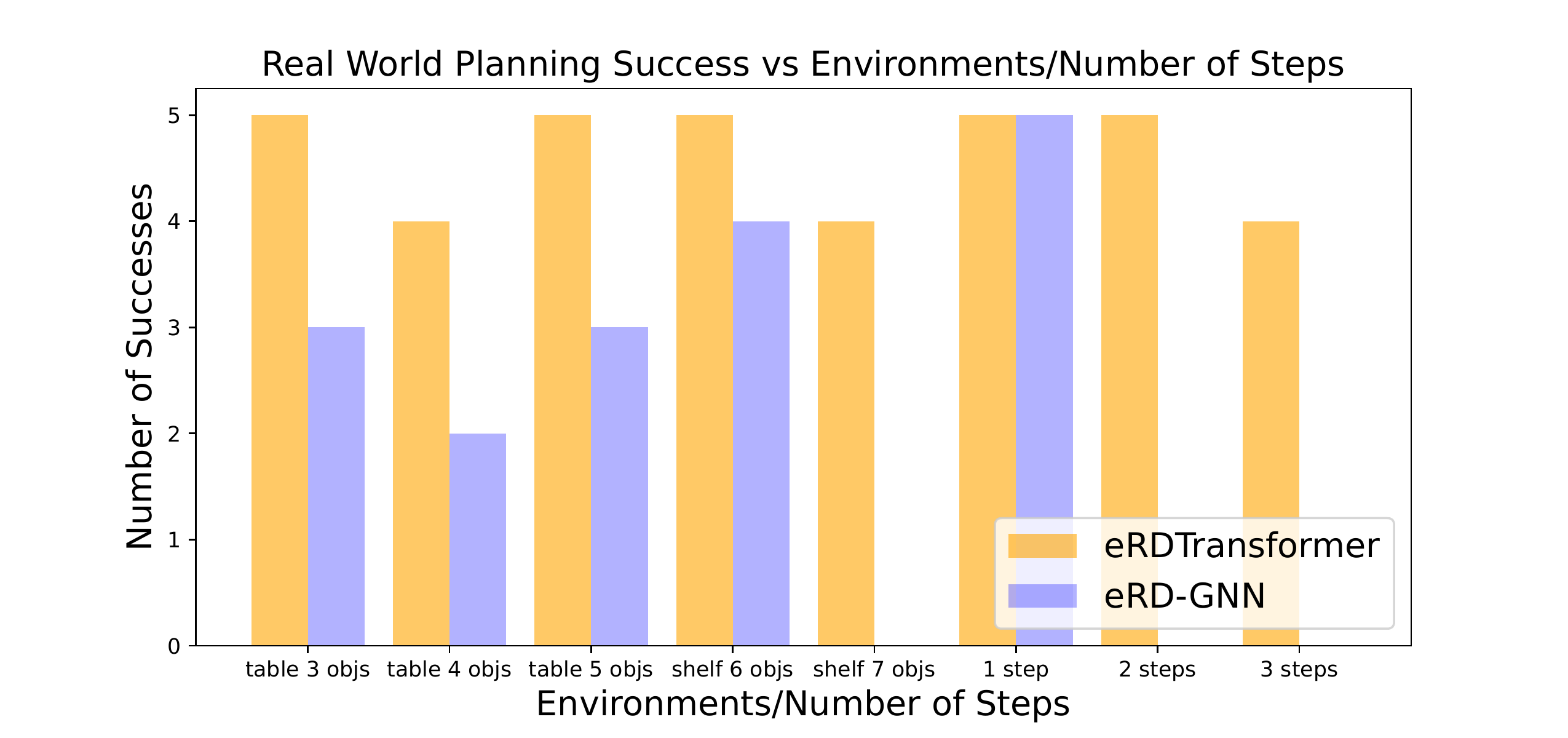} 
    \caption{Real world planning success rate for manipulating YCB objects for single-step and multi-step planning. We show the comparison between \etedyn{} and \egnn{}. Empty bars denote 0 success. We ran 5 trials for each task per model.}
    \label{fig:real_all_env}
\end{figure*}

\begin{figure*}[ht!]
    \centering
      \vspace{-10pt}
      \includegraphics[width=1.98\columnwidth,clip,trim=0mm 0mm 0mm 0mm]{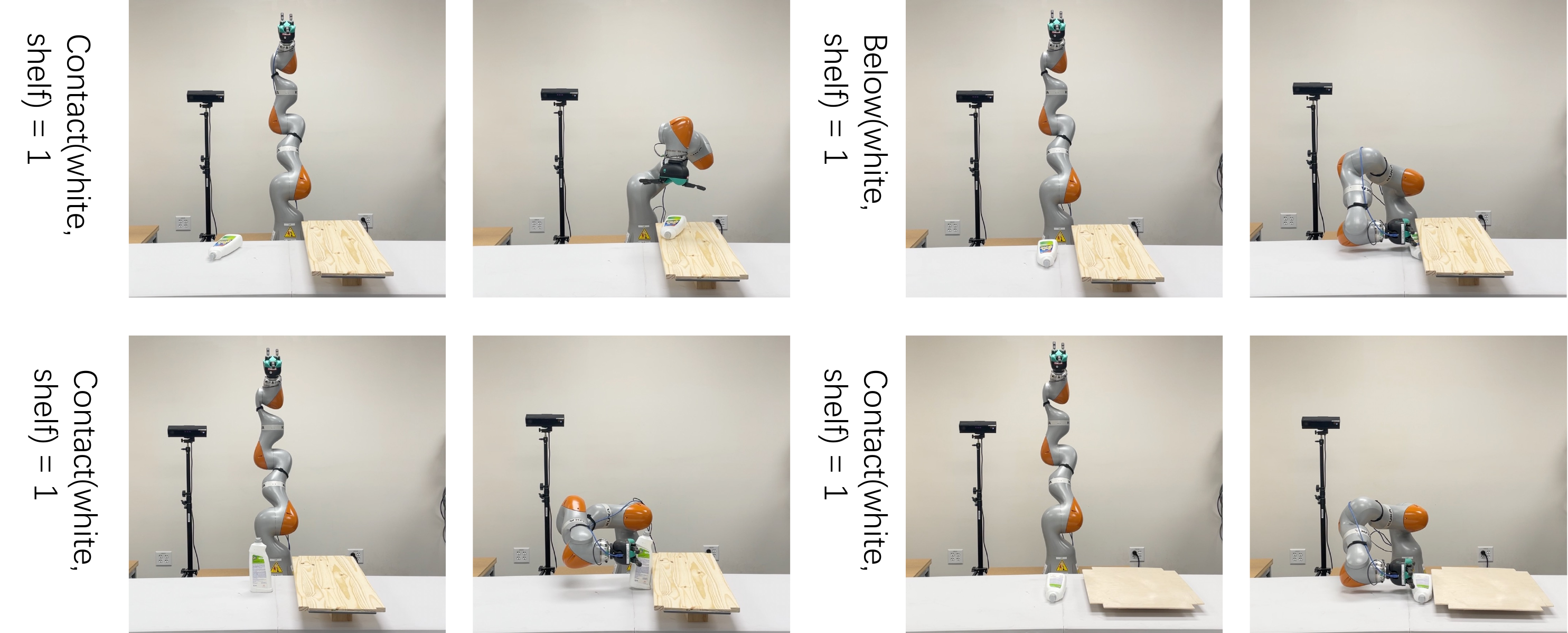}  %
      \caption{We show several real-world examples of how our framework can enable planning for multiple objects across varying shelf-like environments. We show that we can reason about different objects and shelf poses, shape, and height; while also handling different goal relations.
      }
\label{fig:common_sense_1}
\end{figure*}

\begin{table}[ht!]
\centering
\caption{Different F1 score to evaluate object-environment \\ relational dynamics}
\begin{tabular}{ p{1.8cm}p{1.8cm}p{1.5cm}p{1.8cm}p{1.5cm}}
 \hline
 Approaches & Prediction & Detection & Env Detection\\
 \hline
 eRDTransformer   & 0.917 & 0.959 & 1.000 \\
 eRDT-trans-dec  & 0.846 & 0.950 &  1.000 \\
 eRD-GNN     & 0.790 & 0.950 & 1.000\\
 RDGNN  & 0.795 & 0.961 & N/A \\
 eRDT-mlp-dyn  & 0.472 & 0.957 & 1.000\\
 \hline
\end{tabular}
\label{table:evaluation_f1_env}
\end{table}

\subsection{The benefit of predicting whether segments are manipulable}
Our final evaluation in simulation shows the benefit of predicting whether or not segments are manipulable improves planning performance. 
Fig.~\ref{fig:Env_identity} shows that not only does pruning immobile segments from consideration improve planning efficiency, it also improves the success rate. We can attribute this to the planner sometimes electing to manipulate a table or shelf instead of the relevant objects.

\begin{figure*}[ht]
    \centering
      \vspace{-10pt}
      \includegraphics[width=1.98\columnwidth,clip,trim=0mm 0mm 0mm 0mm]{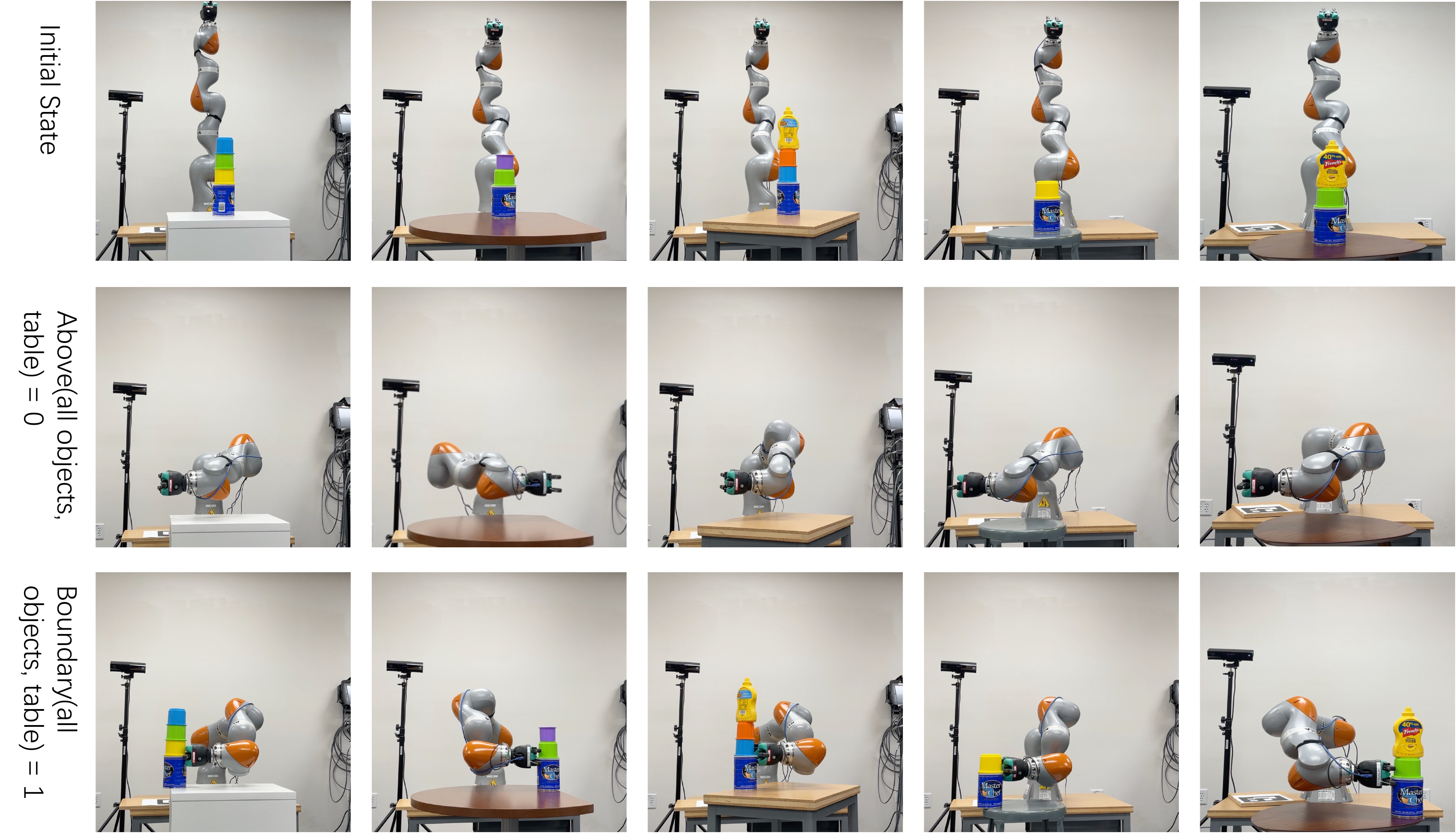}  %
      \caption{Experiments visualizing the nuanced reasoning of our learned model. Given the same initial scene the robot is tasked with moving all objects either to the boundary or off of the supporting table. The robot succeeds for each of five tables of varying shape, size, and height. These results highlight the model's ability to ground the object-environment semantic concepts to the geometry of the observed scene.} \label{fig:real_visulization}
\end{figure*}

\subsection{Real World Manipulation Evaluation}
We now evaluate the manipulation success of \etedyn{} and the best performing GNN model \egnn{} in the real world. 
In particular we examine to what extent the transformer models trained purely in simulation
only on cuboid objects across varied environments transfer to real world YCB objects~\cite{calli2015ycb} and environments without any fine-tuning.
We first conducted quantitative experiments by executing 5 trials for each model on different environments and different number of steps. 
Fig.~\ref{fig:real_all_env} shows that the \etedyn{} once again performs consistently better than \egnn{}.
We then compare multi-step experimental results in the real world as shown in Fig.~\ref{fig:real_all_env}.
We find again that \etedyn{} performs best up to three steps while \egnn{} performs very poorly with more than 1 step. 
Note we only train with random-sized cuboid objects in simulation but can generalize to YCB objects in the real world without any fine-tuning.

\subsection{Discussion}

We first show the generalization ability of our framework in terms of different numbers of objects, different shapes of objects, and varied environments. We show some training examples with varied environments in Fig.~\ref{fig:visualization_environments}. For the table environment, all training examples contain three cuboid objects of varying size above a rectangular table of varying size and orientation. During testing, we test with 2--4 YCB objects above the table and 5 different table shapes, shown in Fig.~\ref{fig:real_visulization}.

For the shelf environments, we train with four cuboids in the environment. We test on two different real-world shelves shown in Fig.~\ref{fig:real_bookshelf} and different simulated shelves in Fig.~\ref{fig:simulation_bookshelf}. Our tests examine generalization to 4--6 YCB objects in the environment shown in Fig. 1 and Fig.~\ref{fig:real_bookshelf}. 
Furthermore, even though our training dataset contains two layers of shelves, our framework can generalize to complex environments with 4 or 6 layers of shelves of varying heights viewed from novel view points as shown in Fig.~\ref{fig:simulation_bookshelf}.

We qualitatively show the capability of \etedyn{} to reason about the geometry of varied environments in Fig.~\ref{fig:common_sense_1}.
In particular these results highlight the understanding of the subtle difference between environments encoded in the learned model. For the same scene the robot understands how to manipulate an object to be above, under, or in contact with a shelf. 
Furthermore, given the same goal relation \texttt{contact(white cleaner, shelf) = 1} with the same environment, but different initial object pose, standing versus lying down, our framework can choose between different actions, picking versus pushing, to achieve the goal relations. The robot can also choose to use pick-and-place to achieve a desired object-environment contact relation when the shelf is high and chooses to push when the shelf is low.
 
These evaluations show the ability of the network to reason about semantically different outcomes of geometrically similar actions and relative object poses.
We find it particularly compelling that our approach can reliably reason about geometric implications between objects and environment structures of varying shape and pose. As shown in Fig.~\ref{fig:real_visulization} the model learns to differentiate between pushing a collection of objects to the boundary of versus pushing off of tables of different geometry. Further the model can be used to reliably plan to one or the other, selecting pushing actions with very similar parameters that generate significantly different outcomes.

Finally, we note that during early experiments with smaller training datasets, we found that our GNN-based model performed comparably to the transformer-based networks.
However, in the presence of larger, more diverse datasets, the transformer-based models perform significantly better than the GNN-based approaches.
Our insight is that using a transformer-based model to encode latent dynamics is the core component to increase the model's capability to handle larger and more diverse datasets.

Finally, we note that during the evaluation of inter-object interactions shown in Fig.~\ref{fig:planning_success_comparison_no_env} and table~\ref{table:evaluation_f1_no_env}, \etedyn{} performs comparable to \rdgnn{}. However, during a more complex evaluation of object-environment interactions shown in Fig.~\ref{fig:planning_success_comparison_env} and table~\ref{table:evaluation_f1_env}, \etedyn{} consistently outperforms the GNN-based approaches.

As shown in our ablation study~(Sec.~\ref{sec:abalation_study}), using the transformer as the dynamics model provides the most significant benefit over alternative models. 
We attribute the transformer-based model's capability in larger scenes with more complex inter-object and object-environment interactions to the use of the learnable attention metrics. 
Specifically, the attention mechanism in the transformer-based dynamics enables the model to learn not just how a specific action affects objects but which objects and parts of the environment are affected. 
This is in contrast to the generic GNN's message passing layers which must pass information through multiple layers to ensure information about all objects passes to all others. 
We note that transformers can be understood as a specific implementation of a GNN using self-attention as the message passing protocol.

\section{Limitations and Future Work}\label{sec:limitation}
Our framework has several limitations.
First, we do not perform replanning during multi-step robot executions, which requires high accuracy of the latent dynamics prediction to succeed at multi-step planning. As such, we only show up to 3 steps for multi-step planning. This connects with our rather limited graph search, which was sufficient for these experiments, but we know would scale inefficiently to longer-horizon and more complex tasks. 
This motivates us to integrate better task planners into our framework. 
Our work is the first one to learn the relational dynamics for multi-object-environment with partial-view point clouds. We learn the environment identity as an easy precondition, but this is much simpler than most state-of-the-art TAMP solvers. 
We think learning a more complete and expressive set of preconditions could enable integrating PDDL~\cite{garrett2020pddlstream} into our learned relational dynamics. This in turn would enable the framework to generalize to longer robot manipulation tasks. However, we note that our experiments show a similar trend in planning success rate and prediction F1 score. This indicates our planning algorithm works effectively when given an accurate predictor. Thus, anyone looking to further improve performance should first focus on improving prediction accuracy.

Second, we don't explicitly integrate the robot's reachable space into our planning framework, which causes the output skill primitive from our approach to sometimes not be executable on the robot. 
We use the RRTConnect motion planner in~\cite{coleman2014reducing} to execute the skill in our paper, but this motion planner does not have a high success rate especially in complex environments such as those involving the shelf. We augment the generic planning by including offset reaching and retraction waypoints to reduce the burden on the motion planner. We plan to incorporate the more advanced motion planner proposed in~\cite{sundaralingam2023curobo} to improve the motion planner success rate.

Third, on a theoretical level we have no proof that the relations we use in training our representation provide a sufficient basis for predicting all inter-object interactions of interest. Currently, we only have empirical results to show they seem to work well. 
For example, we don't include an \texttt{inside} relation in this paper so we cannot plan to achieve goals like ``put the apple inside the basket" or ``put the sugar box inside the cabinet". 
As an additional training issue, we only train with block-shaped objects. While this proved sufficient for demonstrating the benefits of relational dynamics, we do not capture detailed shape information that robots must reason about for more complicated tasks and interactions.

Furthermore, our framework cannot reason about any occluded objects nor any novel objects that appear after a skill execution. Said differently, all relevant objects must be observed in the initial scene. 
This limits our work's applications to more complex environments such as households or offices where objects are inside cabinets or boxes. 
To address this issue, we wish to incorporate object tracking and memory into our model to enable replanning and reasoning about objects that become occluded or disoccluded during manipulation. 
To work with more complex environments, we also need to include more low-level skills like opening/closing a cabinet or pouring a cup of liquid. 
Another interesting direction for future work would be to integrate natural language goals~\cite{liu-icra2022-structformer, liu2022instruction, brohan2022rt}.

\section{Conclusion}\label{sec:discussion}
We propose the first TAMP framework with learned, multi-object relational dynamics based on input from partial-view point clouds. 
Our proposed novel framework can reason about relations between unknown objects and varied environments.
By encoding how actions change the relations between objects and the environment, our approach can achieve multi-step planning with a variable number of objects and environmental components.
We demonstrate the ability of our model to understand how subtle geometric change in different environments effect logical relations.
Through large-scale experiments in simulation and the real world, we show the effectiveness of our approach in manipulation planning.
We can attribute this to better accuracy when predicting both object-environment relations and latent space dynamics compared to baselines analogous to previously proposed approaches.
In particular, our proposed transformer-based model performs best in our evaluations.
Further we showed relations provide a better source of supervision for training our model for planning than using a pose estimation loss.

Overall, our approach provides the first example of predicting manipulation sequences using learned relational dynamics from partial view point clouds. 
We leverage these predictions for planning and executing dynamic rearrangements with multiple objects and environments on a physical robot.

\section*{Acknowledgments}
The authors thank Mohit Sharma, Chris Paxton, and Mohanraj Devendran Shanthi for useful discussion.
This work was partially supported by NSF Award \#2024778, by DARPA under grant \mbox{N66001-19-2-4035}, and by a Sloan Research Fellowship. 

\bibliographystyle{IEEEtran}
\bibliography{references}

\end{document}